\documentclass[11pt]{article}

\usepackage[final]{acl}

\usepackage{times}
\usepackage{latexsym}
\usepackage{amsmath}
\usepackage{booktabs}
\usepackage{tabularx}
\usepackage{soul}
\usepackage{xcolor} % 可选：自定义高亮颜色
\sethlcolor{yellow!30}
\usepackage{subcaption}
\usepackage{booktabs}
\usepackage{array}
\usepackage{makecell}
\usepackage{longtable}
\usepackage{hyperref}

\usepackage[T1]{fontenc}
\usepackage[utf8]{inputenc}

\usepackage{microtype}

\usepackage{inconsolata}

\usepackage{graphicx}

\usepackage{enumitem}

\title{PersonaArena: Dynamic Simulation for Evaluating and Enhancing Persona-Level Role-Playing in Large Language Models}

\author{
 \textbf{Wenlong Shi\textsuperscript{1}},
 \textbf{Jianxun Lian\textsuperscript{2 *}},
 \textbf{Mingqi Wu\textsuperscript{3}},
 \textbf{Haiming Qin\textsuperscript{1}},
 \textbf{Mingyang Zhou\textsuperscript{1}},
 \\
 \textbf{Xing Xie\textsuperscript{2}},
 \textbf{Naipeng Chao\textsuperscript{4,5}},
 \textbf{Hao Liao\textsuperscript{1,5 *}},
\\
 \textsuperscript{1}College of Computer Science and Software Engineering, Shenzhen University, China
 \\
 \textsuperscript{2}Microsoft Research Asia
 \\
 \textsuperscript{3}Microsoft Gaming
\\
\textsuperscript{4}School of Media and Communication, Shenzhen Univerisity
\\
\textsuperscript{5}Provincial Key Laboratory of Intelligent Communication and Digital Society Governance, Shenzhen University
\\
{2553103006@mails.szu.edu.cn},
{jianxun.lian@outlook.com},
{haoliao@szu.edu.cn}
\\
\\
}

\begin{document}
\maketitle
\renewcommand\thefootnote{\fnsymbol{footnote}}
\footnotetext[1]{Corresponding authors.}
\begin{abstract}
Large language models (LLMs) increasingly serve as interactive social agents, yet their ability to maintain coherent and authentic persona-level role-playing remains limited, particularly in realistic social scenarios. Existing research predominantly focuses on character-level settings and relies on static evaluation formats, failing to capture the complexity of everyday social interactions. In this work, we present \textbf{PersonaArena}, a dynamic simulation framework for evaluating and improving persona-level role-playing in LLMs. PersonaArena leverages a large, filtered corpus of user-generated social content to construct a nuanced persona bank, and elicits multi-turn, context-rich interactions within simulated social environments. Our framework features a multi-agent debating judge for holistic and unbiased assessment. Through extensive experiments, we demonstrate that PersonaArena enables rigorous evaluation and enhancement of LLMs' role-playing capabilities, advancing the development of more authentic and socially adept AI agents. The code for the PersonaArena framework is available at our public GitHub repository: \url{https://aka.ms/personaarena}.
\end{abstract}

\section{Introduction}
Large language models (LLMs) are increasingly deployed as interactive agents ranging from social companions to virtual simulations. A key to their effectiveness is role-playing: adopting a persona and maintaining coherent behavior over time. Such capabilities yield more engaging, personalized, and believable interactions, which are crucial for sustained user engagement and more lifelike social presence. Yet despite advances, LLMs — especially those of moderate size (e.g. ~8B parameters) — often lag in persona fidelity, consistency, and adaptability.~\citep{zhou2025personaeval,chen2024socialbench,samuel2024personagym}. These gaps highlight the need for rigorous evaluation and methods specifically aimed at improving persona-level role-playing in order to realize more authentic AI agents.

Majority of role-playing literature focuses on the character-level setting, in which roles correspond to well-known characters from novels, films, scripts, or celebrities~\cite{wang2025coser,wang-etal-2024-rolellm,tu2024charactereval,wang-etal-2025-characterbox,li2023chatharuhi,shao-etal-2023-character,chen2022large}. However, those characters are often part of popular culture and thus may function more like commonsense knowledge that LLMs memorize rather than truly reason about. Moreover, such characters are frequently exaggerated or idealized, deviating significantly from ordinary human behavior. Consequently, strong performance on character-level role-playing does not guarantee reliable simulation of everyday social interactions — the kind of behavior foundational for AI in social science~\cite{aher2023using,hewitt2024predicting}. 

In this paper, we focus on persona-level role-playing. Compared with character-level settings, research in this direction remains relatively limited, with representative works including~\cite{zhang-etal-2018-personalizing,jandaghi2023faithful,wang2025opencharacter,peng2024quantifying,samuel2024personagym}. However, several challenges remain unresolved.
(1) Limited Faithfulness of human-written dialogues. Early datasets such as Persona-Chat~\cite{zhang-etal-2018-personalizing} collect conversations from crowdsourced workers asked to role-play assigned profiles. Yet unlike professional actors, most workers struggle to authentically simulate others' thoughts and behaviors. Ideally, only workers would converse as themselves to ensure authenticity.
(2) Limited evaluation metrics. Existing studies often rely on surface-level measures such as hit@k, perplexity, or BLEU~\cite{zhang-etal-2018-personalizing,jandaghi2023faithful}, or narrowly focus on specific aspects such as faithfulness~\cite{peng2024quantifying} or identity recognition~\cite{zhou2025personaeval}, leaving broader aspects of persona consistency and adaptability underexplored.
(3) Restricted interaction formats. Role-playing behavior is frequently elicited through self-report question-answer pairs~\cite{samuel2024personagym,klinkert2024evaluating}, which diverge from realistic open-ended conversational scenarios where persona expression naturally unfolds over context.

To address the above challenges, we propose PersonaArena, a dynamic simulation framework for evaluating and enhancing persona-level role-playing in LLMs. We observe that massive user-generated content on social platforms, such as blog posts, naturally conveys individuals' personas and social experiences.{Building on a raw dataset containing over 19k users and 681k posts, we perform quality filtering and construct a persona bank of 1k distinct profiles.} This bank captures nuanced and diverse social identities that extend beyond simple demographic descriptors.
Rather than relying on static persona-based Q\&A probing, PersonaArena introduces a social simulation framework designed to elicit multi-turn, context-rich interactions resembling realistic social exchanges. The framework consists of two core components: an environment agent, which coordinates scenario development and tracks evolving interactions, and non-player characters (NPCs) that engage with the protagonist agent (the LLM under evaluation). To ensure fair and comprehensive assessment, we further design a multi-agent debating judge that evaluates persona fidelity, coherence, and adaptability.
Through extensive experiments, we show that PersonaArena effectively elicits high-quality behavioral trajectories, enabling faithful evaluation of role-playing capabilities. Moreover, data generated within PersonaArena can be leveraged as post-training material to further improve LLMs' persona consistency and realism.

Main contributions are summarized as follows:
\begin{itemize}[leftmargin=*]
\item We propose PersonaArena, a social simulation framework that elicits persona-based behaviors through  dynamic multi-turn interactions.
\item We introduce a multi-agent debating judge to evaluate role-playing quality in a holistic and unbiased manner.
\item We conduct extensive experiments demonstrating the effectiveness of PersonaArena in evaluating LLMs' role-playing capabilities. 
\item We show that data elicited within PersonaArena can further enhance role-playing performance through targeted post-training.
\end{itemize}

\section{The Simulation Framework}

\label{sec:framework}

% \subsection{Overview}

\begin{figure*}[t]
\centering
\includegraphics[width=\linewidth]{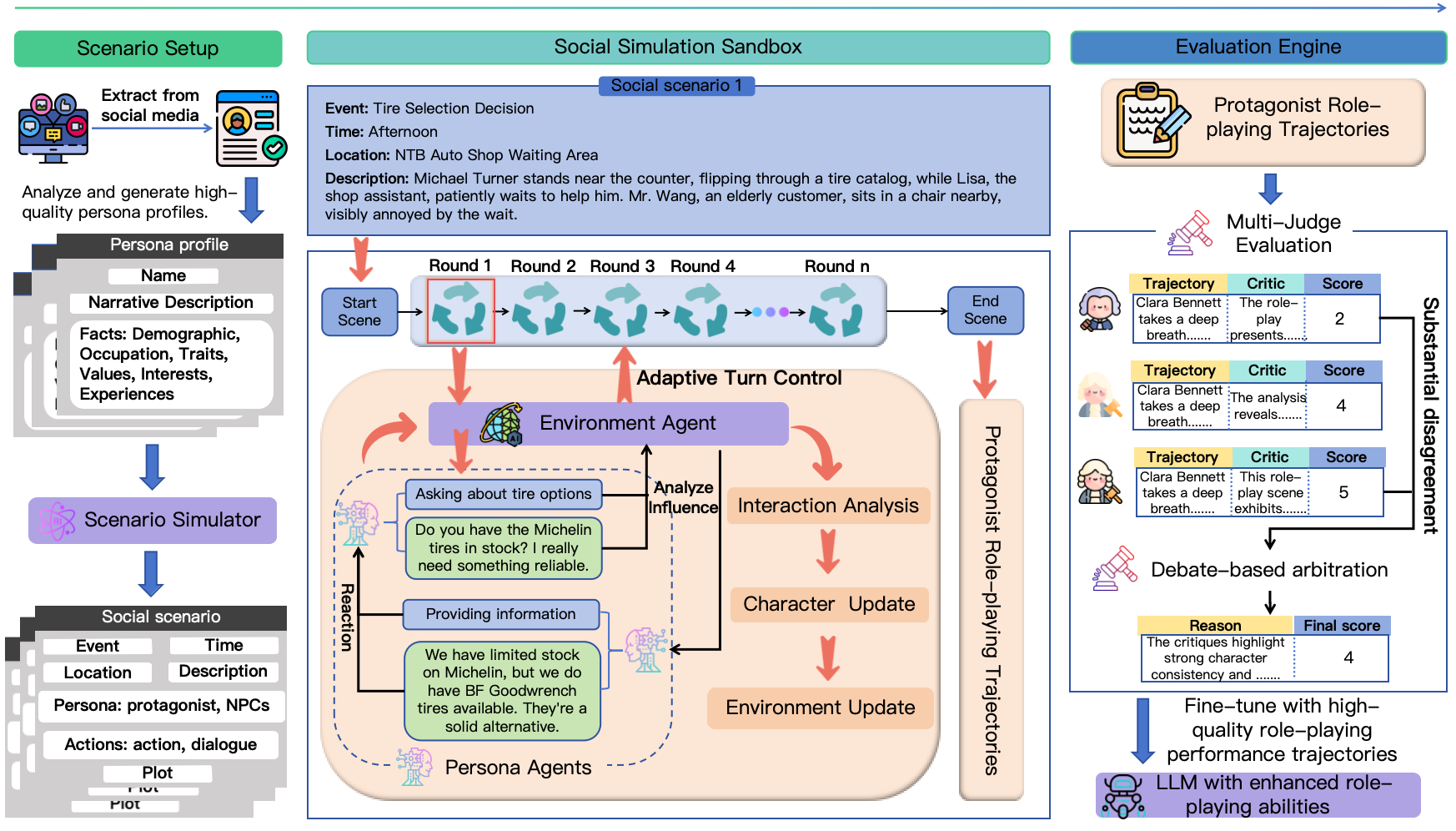}
\caption{Overall architecture of \textbf{PersonaArena}.  
    Persona-level profiles are instantiated into autonomous character agents interacting within dynamically generated social scenarios.  
    The Environment Agent monitors the coverage of five persona dimensions and triggers early stopping upon sufficient expression.}
\label{fig:personaArena}
\end{figure*}

We introduce \textbf{PersonaArena}, a text-based interactive evaluation framework for assessing large language models (LLMs) on their persona-based role-playing abilities within multi-agent social contexts. Building on prior work in interactive virtual world frameworks~\cite{park2023generative,wang-etal-2025-characterbox}, PersonaArena generates dynamic, persona-grounded simulations where multiple agents interact and respond to evolving social environments. Each evaluation scenario $\mathcal{A}$ comprises three key components: $\mathcal{A} = (\mathcal{P}, \mathcal{S}, \mathcal{E})$, where $\mathcal{P}$ is the set of personas, $\mathcal{S}$ denotes the interactive scenario, and $\mathcal{E}$ is the evaluation engine. The overall evaluation process consists of three stages: (\textit{i}) \textbf{Scenario Setup}, (\textit{ii}) \textbf{Social Simulation in a Sandbox Environment}, and (\textit{iii}) \textbf{Evaluation via Multi-Agent Debates}. Fig.~\ref{fig:personaArena} illustrates the overall process.

\subsection{Scenario Setup}
\label{sec:scenario_setup}
\paragraph{Persona Bank.}
We argue that persona settings limited to basic demographic information are insufficient for rigorous and realistic role-play evaluation. To overcome this limitation, we utilize the rich and nuanced life experiences that individuals share on social platforms such as blogs. Specifically, we construct a diverse \textit{persona corpus} from user-generated blog posts in the publicly available Blog Authorship dataset\footnote{\url{https://huggingface.co/datasets/barilan/blog_authorship_corpus}}. An LLM is employed to preprocess the raw data by replacing private information (e.g., names, emails, home addresses) with randomly generated substitutes and to infer comprehensive persona profiles, incorporating demographic, occupational, and psychological attributes such as values. The resulting corpus spans a broad spectrum of personas, enabling more authentic and challenging role-playing objectives for social simulation. Each persona $p_i$ comprises a narrative description and a structured set of factual attributes, represented as $p_i = \{ \text{name}, \text{narrative}, \text{facts} = (d, o, \pi, v, I, e) \}$, where $d$ denotes demographic information, $o$ occupation, $\pi$ personality traits, $v$ values, $I$ interests, and $e$ experiences. The final persona corpus contains 1,000 unique personas.

\paragraph{Scenario Setup.}
Given a target persona $p_i$, the \textit{Environment Agent} automatically constructs a realistic social scenario $s_i$ that reflects the persona's characteristics and narrative background.  
Each scenario consists of a textual event description, temporal and spatial context, a protagonist, and two to three supporting NPCs.  
Character descriptions are further enriched with factual priors extracted from $(o, \pi, v, I, e)$ to ensure semantic coherence between persona definition and situational context.  
This module establishes the initial conditions for controlled and comparable multi-agent interactions in subsequent simulations.

\subsection{Social Simulation Sandbox}

\paragraph{Persona Agents.}
Following \cite{wang-etal-2025-characterbox}, we instantiate each persona as an autonomous \textit{Character Agent} with goal-conditioned reasoning and a Belief--Desire--Intention (BDI) structure~\cite{georgeff1998belief}, enabling agents to maintain evolving internal states throughout social interactions.  
There are two types of character agents:
\begin{itemize}[leftmargin=*]
    \item \textbf{Protagonist Agent:}  
    Protagonist is controlled by the LLM to be evaluated, serves as the central focus of interaction.  
    In each simulation round, it observes the current environment, retrieves relevant episodic memories, and generates goal-directed responses that remain causally coherent with prior events.  
    Its internal memory module is implemented as a vector-based semantic retriever, allowing persistent recall of contextual cues and emotional traces.  
    Within the BDI framework, the protagonist dynamically updates both Self-Belief (self-awareness of identity, motivation, and intent) and Env-Belief (understanding of others and situational context), reflecting adaptive reasoning and social cognition.

    \item \textbf{NPC Agents:}  
    The supporting NPCs are instantiated from a fixed set of high-capability LLMs (e.g., Qwen3-32B or GPT-4 models) to ensure stable, coherent behavior across simulations. Unlike the protagonist, these agents follow reactive policies guided by predefined personas and dialogue constraints. Under the same BDI formulation, NPCs mainly adjust their Env-Belief to remain behaviorally consistent with the protagonist’s evolving actions, while keeping Self-Belief static to preserve character identity and narrative continuity. This asymmetric design enables PersonaArena to isolate the reasoning and adaptability of the evaluated model while maintaining controlled, reproducible multi-agent interactions.
    \end{itemize}

To ensure clarity and interpretability, intermediate reasoning traces (e.g., hidden chain-of-thoughts or system annotations) are automatically filtered, leaving only the final utterances that contribute to the narrative flow.

\begin{figure}[t]
\centering
\includegraphics[width=\linewidth]{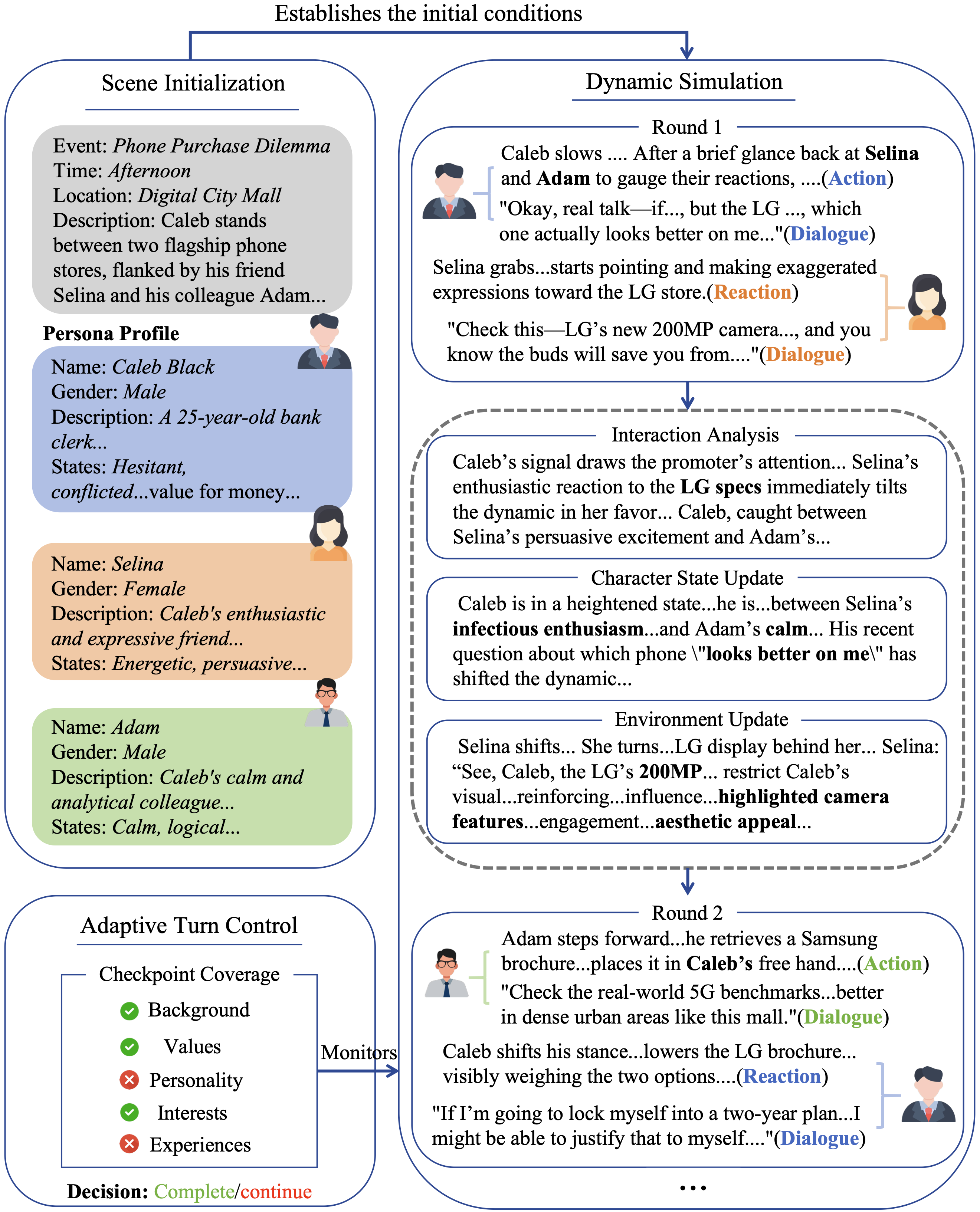}
\caption{Dynamic simulation in PersonaArena. An example interaction loop is shown, where scene initialization defines the protagonist, NPCs, and event context, and subsequent rounds proceed through action–reaction exchanges under turn-level control. The figure highlights interaction analysis, adaptive turn control, character update, and environment update for maintaining coherent and persona-consistent trajectories.}
\label{fig:DS}
\end{figure}

\paragraph{Environment Agent.}
The Environment Agent acts as the global controller that orchestrates all role interactions, monitors progress, and maintains overall narrative consistency.  
Its main functions include the following:
\begin{itemize}[leftmargin=*]
    \item \textbf{Interaction Analysis:}  
    When a character $c_i$ takes an action or utters a statement, the Environment Agent evaluates its potential influence on other participants by considering their current physical and psychological states.  
    It then identifies the character $c_r$ most affected and likely to respond, and passes the action $a_i$ and its derived impact $f_r$ to $c_r$.  
    The responding agent $c_r$ generates a reaction accordingly, after which the Environment Agent summarizes the interaction outcome as $R$, updating shared context such as memory traces, spatial positions, and emotional states.

    \item \textbf{Adaptive Turn Control:}  
    After each dialogue round, the Environment Agent performs checkpoint-based monitoring to verify whether the protagonist has sufficiently expressed persona-relevant information across key semantic dimensions.  
    A five-dimensional checkpoint set is defined as  
    $\mathcal{C} = \{\text{Background}, \text{Personality}, \text{Values}, \text{Interests},$ $ \text{Experiences}\}$.  
    For each dimension, the agent examines whether clear evidence has emerged in the protagonist’s dialogue and action history.  
    Evidence is accumulated across rounds, and a multi-criteria early-stopping rule is applied to balance expressiveness and efficiency.  
    The interaction terminates naturally once most or all dimensions are adequately covered or when incremental evidence becomes redundant.  
    This mechanism ensures comprehensive yet concise persona coverage, forming the basis for fair and consistent evaluation.

    \item \textbf{Character State Update:}  
    The Environment Agent updates the internal states of both the protagonist and NPCs based on their own actions and interaction results.  
    If a character $c_i$ receives a response from others, its state update integrates both self-action and external feedback; otherwise, it is updated solely according to its own behavior.  
    This design maintains temporal and emotional continuity throughout the multi-agent simulation.

    \item \textbf{Environment Update:}  
    At the end of each round, the Environment Agent synchronizes the global environment $\mathcal{E}$ based on the cumulative effects of agents' actions.  
    If no action explicitly affects environmental variables, the environment remains unchanged.  
    This process maintains causal consistency between agent behaviors and environmental dynamics, ensuring coherent world evolution across the simulation.
\end{itemize}
The process of scene initialization and dynamic simulation is shown in Fig.~\ref{fig:DS}

\subsection{Evaluation Engine}

\paragraph{Metrics.}
Upon completion of the simulation, PersonaArena evaluates the resulting trajectory using a comprehensive suite of eight metrics, each designed to capture complementary aspects of role-playing performance. The first seven metrics, including Knowledge Accuracy (KA), Behavioral Accuracy (BA), Emotional Expression (EE), Personality Traits (PT), Immersion (IM), Behavioral Coherence (BC), and Adaptability (AD), are adopted from \cite{wang-etal-2025-characterbox}. To further assess the novelty and diversity of elicited behaviors, we introduce a new metric, \textit{Interaction Richness (IR)}, which measures the breadth and depth of social responses, thereby reflecting the creativity and variety of role-play performance. Detailed definitions of all eight metrics are provided in Appendix~\ref{sec:app_metrics}.

\paragraph{Multi-Judge Evaluation.}
To ensure fairness, robustness, and interpretability, PersonaArena employs $K$ independent LLM judges $\mathcal{J}=\{J_1,\dots,J_K\}$ to evaluate each simulation.  
Each judge independently scores all eight evaluation dimensions, producing a vector of quantitative ratings.  
Final scores are obtained through mean aggregation:
\begin{equation}
\bar{\mathbf{s}}_i = \frac{1}{K} \sum_{J_k \in \mathcal{J}} \mathbf{s}_i^{(J_k)},
\label{eq:mean_score}
\end{equation}
and inter-judge agreement is continuously monitored to validate scoring reliability.

\textbf{Debate-based arbitration.}  
When substantial disagreement is detected among judges, PersonaArena initiates a debate-based arbitration process to reconcile inconsistent evaluations.  
In this stage, each disagreeing judge $J_k$ is required to submit its own \emph{critic statement}, including (1) its rating $\mathbf{s}_i^{(J_k)}$, (2) textual justification, and (3) evidential excerpts supporting the assigned score.  
A designated referee model $R$ then reviews all submitted critic statements and evidence, synthesizes them into a unified rationale, and issues a reconciled score $\mathbf{s}_i^{(R)}$ with an explicit explanation.

The reconciled rationale and score are then adopted as the final result for the disputed metric, whereas the scores of all non-disputed metrics are still obtained through the mean aggregation in Eq.~\ref{eq:mean_score}.

\section{Enhancing Role-playing Ability}
\label{sec:Enhancing}

PersonaArena not only generates rich behavior trajectories for role-playing evaluation but also serves as a resource for enhancing the role-playing capabilities of LLMs. To achieve this, we leverage high-quality trajectories collected from multi-agent simulations as fine-tuning data for LLMs.
We explore two fine-tuning paradigms, i.e., \textbf{Supervised Fine-Tuning (SFT)} and \textbf{Direct Preference Optimization (DPO)}~\cite{rafailov2023direct}. 
In SFT, high-quality trajectory samples, comprising character utterances, contextual summaries, and interaction goals, are directly used to fine-tune the base model via supervised learning. This process allows the model to imitate desirable behavioral patterns and narrative coherence observed in expert-generated trajectories.  DPO is a preference-based fine-tuning strategy that explicitly contrasts high-quality and low-quality trajectories. Specifically, for each dialogue context, we construct a trajectory pair based on the evaluated behaviors of different LLMs. The \textbf{high-quality trajectory} corresponds to a response sequence that received a higher evaluation score, while the \textbf{low-quality trajectory} corresponds to one with a lower score. The DPO objective encourages the model to assign higher likelihoods to high-quality trajectories while penalizing those that are less aligned with desired behavioral patterns. Unlike reinforcement learning methods such as PPO~\cite{bai2022training}, DPO achieves direct preference optimization without requiring reward modeling or rollout sampling, ensuring both simplicity and stability in implementation.

\begin{table*}[t]
\centering
\small
\setlength{\tabcolsep}{3pt}
\caption{Evaluation results. Each value is presented as mean $_{\text{\scriptsize $\pm$ standard deviation}}$, \textbf{Bold} values indicate the highest scores, and \underline{underlined} values indicate the second-highest scores.}
\label{tab:overall}
\begin{tabular}{lccccccccc}
\toprule
\textbf{Model} & \textbf{KA} & \textbf{BA} & \textbf{EE} & \textbf{PT} & \textbf{IM} & \textbf{BC} & \textbf{AD} & \textbf{IR} & \textbf{Average} \\
\midrule

Phi4 & $3.833_{\pm.08}$ & $3.300_{\pm.08}$ & $3.333_{\pm.08}$ & $3.667_{\pm.12}$ & $3.553_{\pm.05}$ & $3.307_{\pm.17}$ & $3.707_{\pm.12}$ & $3.207_{\pm.07}$ & $3.488_{\pm.08}$ \\
Mistral-small3.2 & $4.140_{\pm.39}$ & $3.667_{\pm.04}$ & $3.680_{\pm.10}$ & $3.940_{\pm.10}$ & $3.787_{\pm.10}$ & $3.553_{\pm.12}$ & $3.773_{\pm.22}$ & $3.487_{\pm.07}$ & $3.753_{\pm.11}$ \\
Grok-3      & $3.973_{\pm.09}$ & $3.520_{\pm.05}$ & $3.493_{\pm.21}$ & $3.867_{\pm.06}$ & $3.567_{\pm.17}$ & $3.567_{\pm.06}$ & $3.687_{\pm.07}$ & $3.333_{\pm.27}$ & $3.626_{\pm.10}$ \\
Llama3.1-8B & $3.847_{\pm.07}$ & $3.307_{\pm.11}$ & $3.220_{\pm.06}$ & $3.567_{\pm.06}$ & $3.353_{\pm.07}$ & $3.120_{\pm.08}$ & $3.387_{\pm.08}$ & $2.847_{\pm.05}$ & $3.331_{\pm.05}$ \\
Llama3.2-3B & $3.900_{\pm.07}$ & $3.147_{\pm.05}$ & $3.067_{\pm.08}$ & $3.527_{\pm.08}$ & $3.300_{\pm.03}$ & $3.100_{\pm.07}$ & $3.327_{\pm.03}$ & $2.947_{\pm.04}$ & $3.289_{\pm.05}$ \\
Qwen3-1.7B  & $3.667_{\pm.08}$ & $3.113_{\pm.16}$ & $3.073_{\pm.14}$ & $3.373_{\pm.13}$ & $3.213_{\pm.11}$ & $2.933_{\pm.15}$ & $2.767_{\pm.16}$ & $2.560_{\pm.12}$ & $3.088_{\pm.12}$ \\
Qwen3-4B    & $3.827_{\pm.11}$ & $3.300_{\pm.11}$ & $3.367_{\pm.11}$ & $3.540_{\pm.12}$ & $3.367_{\pm.11}$ & $3.127_{\pm.15}$ & $3.093_{\pm.12}$ & $2.593_{\pm.15}$ & $3.277_{\pm.12}$ \\
Qwen3-8B    & $3.773_{\pm.11}$ & $3.353_{\pm.03}$ & $3.240_{\pm.05}$ & $3.700_{\pm.09}$ & $3.420_{\pm.06}$ & $3.180_{\pm.05}$ & $3.373_{\pm.04}$ & $2.860_{\pm.05}$ & $3.363_{\pm.04}$ \\
Qwen3-14B   & $4.307_{\pm.05}$ & $3.633_{\pm.06}$ & $3.427_{\pm.04}$ & $3.907_{\pm.04}$ & $3.567_{\pm.05}$ & $3.360_{\pm.04}$ & $3.547_{\pm.07}$ & $2.967_{\pm.05}$ & $3.589_{\pm.03}$ \\
Qwen3-32B   & $4.367_{\pm.12}$ & $3.747_{\pm.06}$ & $3.700_{\pm.11}$ & $3.960_{\pm.09}$ & $3.820_{\pm.10}$ & $3.607_{\pm.08}$ & $3.793_{\pm.07}$ & $3.493_{\pm.13}$ & $3.811_{\pm.06}$ \\
Deepseek-r1-8B    & $3.693_{\pm.23}$ & $3.327_{\pm.14}$ & $3.347_{\pm.07}$ & $3.660_{\pm.15}$ & $3.433_{\pm.15}$ & $3.167_{\pm.17}$ & $3.273_{\pm.16}$ & $2.920_{\pm.10}$ & $3.352_{\pm.13}$ \\
Deepseek-V3.2     & $\underline{4.420}_{\pm.11}$ & $3.767_{\pm.07}$ & $\underline{3.727}_{\pm.06}$ & \textbf{4.187}$_{\pm.09}$ & \textbf{3.993}$_{\pm.04}$ & $3.673_{\pm.05}$ & $3.973_{\pm.05}$ & $3.480_{\pm.03}$ & $3.902_{\pm.05}$ \\
GPT-oss           & $4.013_{\pm.06}$ & $3.520_{\pm.16}$ & $3.440_{\pm.11}$ & $3.653_{\pm.16}$ & $3.580_{\pm.10}$ & $3.393_{\pm.12}$ & $3.660_{\pm.11}$ & $3.280_{\pm.13}$ & $3.567_{\pm.10}$ \\
GPT-3.5           & $3.873_{\pm.07}$ & $3.427_{\pm.21}$ & $3.473_{\pm.23}$ & $3.607_{\pm.09}$ & $3.513_{\pm.18}$ & $3.240_{\pm.19}$ & $3.453_{\pm.10}$ & $3.093_{\pm.23}$ & $3.460_{\pm.16}$ \\
GPT-4o-mini       & $4.210_{\pm.21}$ & $3.638_{\pm.10}$ & $3.645_{\pm.08}$ & $3.940_{\pm.12}$ & $3.769_{\pm.13}$ & $3.579_{\pm.11}$ & $3.679_{\pm.14}$ & $3.293_{\pm.13}$ & $3.719_{\pm.12}$ \\
GPT-4o            & $4.327_{\pm.07}$ & $3.620_{\pm.04}$ & $3.527_{\pm.18}$ & $3.993_{\pm.09}$ & $3.800_{\pm.13}$ & $3.620_{\pm.08}$ & $3.700_{\pm.13}$ & $3.333_{\pm.14}$ & $3.740_{\pm.07}$ \\
GPT-4.1           & $4.373_{\pm.25}$ & \textbf{3.867}$_{\pm.12}$ & \textbf{3.800}$_{\pm.14}$ & $4.040_{\pm.14}$ & $\underline{3.967}_{\pm.13}$ & $\underline{3.820}_{\pm.09}$ & $\underline{4.060}_{\pm.19}$ & $\underline{3.660}_{\pm.10}$ & $\underline{3.948}_{\pm.14}$ \\
GPT-5.1 & \textbf{4.427}$_{\pm.16}$ & $\underline{3.853}_{\pm.05}$ & $3.647_{\pm.22}$ & $\underline{4.107}_{\pm.05}$ & \textbf{3.993}$_{\pm.09}$ & \textbf{3.880}$_{\pm.04}$& \textbf{4.080}$_{\pm.20}$ & \textbf{3.713}$_{\pm.11}$ & \textbf{3.963}$_{\pm.04}$ \\

\bottomrule
\end{tabular}
\end{table*}

\section{Experiment}

\subsection{Evaluation Setting}

\textbf{Scene Initialization.} Our extracted persona bank (see Section~\ref{sec:scenario_setup}) consists of 1,000 user profiles, each featuring detailed background information, occupation, personality traits, and life experiences. 
% (refer to Appendix~\ref{} for further details). 
For each benchmarking run, 10 personas are randomly sampled from the bank and instantiated within realistic scenarios using the environment simulator. These scenarios are consistently employed across all evaluated LLMs. 
Additional details on persona-bank construction, anonymization validation, demographic statistics, and scene categories are provided in Appendix~\ref{sec:app_PBI} and Appendix~\ref{sec:app_SCI}. 

\noindent\textbf{LLMs to be evaluated.}   
We evaluate both closed-source and open-source LLMs with varying model sizes. In each simulation, the tested LLM exclusively plays the \textit{protagonist}, while all NPCs and the \textit{Environment Agent} are controlled by a fixed \textbf{Qwen3-32B} model to ensure interaction consistency and fairness across trials. For quantitative evaluation, we employ a multi-judge framework consisting of three independent LLM judges--\textbf{DeepSeek-R1}, \textbf{Qwen3-32B}, and \textbf{Mistral-small3.2}--along with a \textbf{GPT-4o-mini} arbiter that resolves disagreements when inter-judge variance exceeds a predefined threshold. This setup ensures reliable, multi-perspective assessment of the protagonist’s role-playing performance.
For closed-source models, we evaluate \textbf{GPT-5.1}, \textbf{GPT-4.1}, \textbf{GPT-4o}, \textbf{GPT-4o-mini}, \textbf{GPT-3.5}, and \textbf{Grok-3}\footnote{\url{https://learn.microsoft.com/en-us/azure/ai-foundry/openai/overview}}. For open-source models, we include \textbf{GPT-OSS-20B}\footnote{\url{https://huggingface.co/openai/gpt-oss-20b}}, the \textbf{Qwen3} series ranging from 1.7B to 32B parameters~\cite{yang2025qwen3}, \textbf{Mistral-Small-3.2-24B}\footnote{\url{https://huggingface.co/mistralai/Mistral-Small-3.2-24B-Instruct-2506}}, \textbf{Llama3-8B} and \textbf{Llama3.2-3B}\footnote{\url{https://huggingface.co/meta-llama/Llama-3.2-3B}}, \textbf{Phi-4}~\cite{abdin2024phi}, \textbf{Deepseek-R1}~\cite{guo2025deepseek}, and \textbf{Deepseek-V3.2}~\cite{liu2025deepseek}. All open-source LLMs evaluated are instruction-tuned versions to ensure comparable conversational alignment. The detail prompts provided for LLM is in Appendix~\ref{sec:app:prompt}. 

\subsection{Overall Performance}
Table~\ref{tab:overall} presents the overall evaluation results across eight dimensions for various LLMs. Among all evaluated models, GPT-5.1 achieves the highest overall performance, while Deepseek-V3.2 delivers the strongest results among open-source models, with Qwen3-32B emerging as the best-performing model within the Qwen3 series. A clear scaling trend can be observed within the Qwen3 family: as model size increases from 1.7B to 32B, performance generally improves across most evaluation dimensions. This trend supports the reliability of the benchmark. Additionally, comparisons between models from the same provider but different series, such as GPT-3.5 vs. GPT-4o, suggest that although model scale is important, training objectives, optimization strategies, and overall model design also play a substantial role in determining role-playing quality.

Overall, the results indicate that closed-source GPT-4 and GPT-5 family models continue to set the benchmark for role-playing quality. Meanwhile, the Qwen3 series demonstrates a clear scaling pattern and substantial potential among open-source alternatives, while Deepseek-V3.2 shows that strong open-source models can approach the performance of top-tier closed-source systems. These findings underscore the critical role of both model scaling and training methodology in improving the quality of role-playing behaviors in LLMs.

\begin{table*}[t]
\centering
\small
\caption{Pearson correlation coefficients comparing the single-judge baseline and the multi-judge PersonaArena framework against human evaluations. \textbf{Bold} values denote the strongest alignment with human ratings for each metric, highlighting the reliability advantage of multi-judge evaluation.}
\begin{tabular}{lccccccccc}
\toprule
\textbf{Model} & \textbf{KA} & \textbf{BA} & \textbf{EE} & \textbf{PT} & \textbf{IM} & \textbf{BC} & \textbf{AD} & \textbf{IR} & \textbf{Overall} \\
\midrule
PersonaArena (DeepSeek-r1)    & 0.236    & 0.332    & 0.328    & 0.132 & 0.357    & 0.342 & 0.374 & 0.225    & 0.330 \\
PersonaArena (Mistral-small3.2)    & 0.425    & \textbf{0.583}    & 0.482    & 0.432 & 0.573    & \textbf{0.375} & 0.474 & 0.525    & 0.484 \\
PersonaArena (Qwen3-32B)      & 0.636 & 0.571 & 0.535 & 0.401 & \textbf{0.725} & 0.542 & 0.547 & \textbf{0.768} & 0.669 \\
PersonaArena (Multi-judge)   & \textbf{0.689} & 0.545 & \textbf{0.622} & \textbf{0.558} & 0.670 & 0.419 & \textbf{0.649} & 0.631 & \textbf{0.683} \\
\bottomrule
\end{tabular}
\label{tab:validity}
\end{table*}

\subsection{Reliability and Validity}

To validate the reliability of the proposed multi-judge debate evaluation, we compare the aggregated scores produced by \textbf{PersonaArena} with those from individual LLM judges. As shown in Table~\ref{tab:validity}, the collective judge configuration achieves the highest overall alignment with human ratings, yielding an average correlation score of \textbf{0.683}, surpassing both \textbf{Qwen3-32B} (0.669) and \textbf{DeepSeek-R1} (0.330).
Across all eight evaluation dimensions, PersonaArena consistently exhibits stronger agreement with human judgments. In particular, notable gains are observed in \textbf{KA (0.689)}, \textbf{EE (0.622)}, and \textbf{AD (0.649)}, indicating superior sensitivity to knowledge fidelity, emotional realism, and contextual adaptability.
These results confirm that aggregating multiple LLM judges through debate-based arbitration yields more stable and human-aligned evaluations than single-judge baselines, underscoring the robustness of PersonaArena’s scoring mechanism.
Detailed human-evaluation protocol and an additional multi-judge calibration case study are provided in Appendix~\ref{sec:app_he} and Appendix~\ref{sec:app_judge_case}.

\subsection{Enhancing Role-playing in LLMs}

We fine-tune one representative open-source model, Qwen3-8B, using the proposed SFT and DPO paradigms described in Section~\ref{sec:Enhancing}. The fine-tuned models are evaluated on unseen persona scenarios from PersonaArena to assess their generalization in novel interactive contexts, with GPT-4.1 included as a strong reference baseline.

\begin{figure}[t]
\centering
\includegraphics[width=\linewidth]{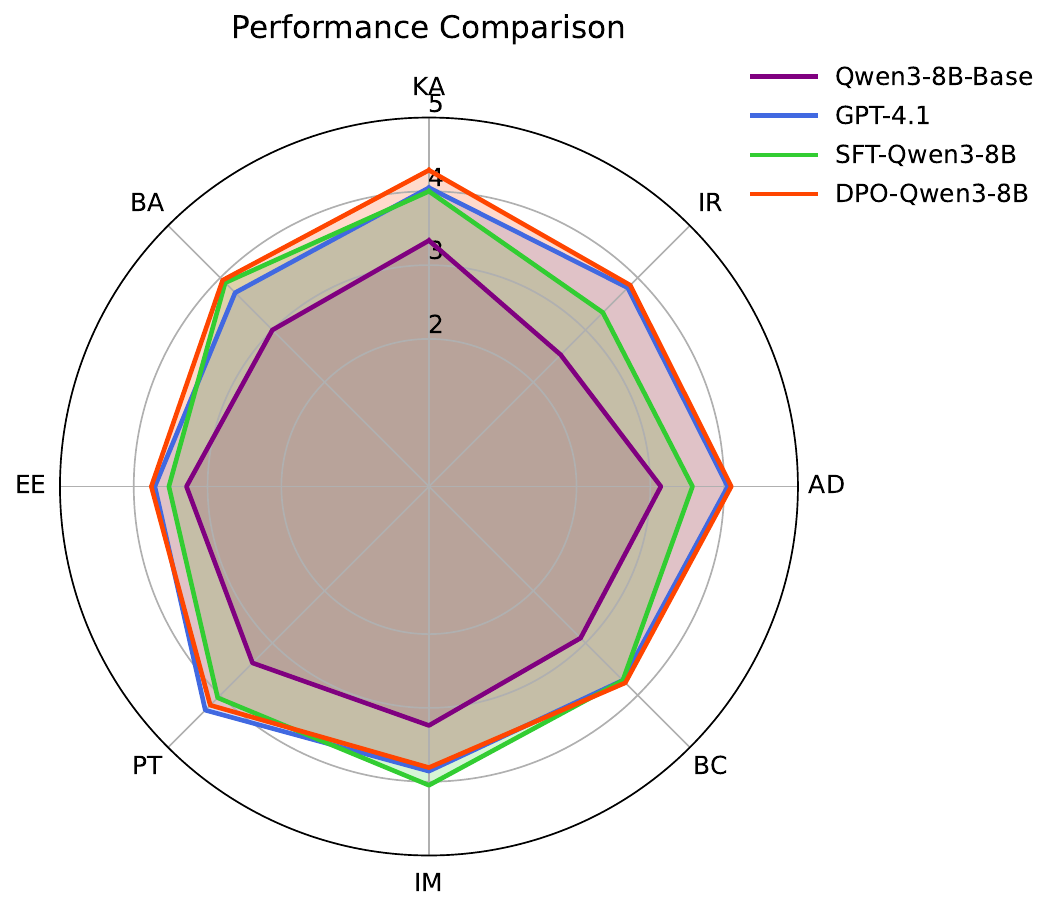}
\caption{Performance comparison between SFT, DPO, and baseline models.}
\label{fig:fine-tune}
\end{figure}

\textbf{SFT fine-tuning.} In this stage, {Qwen3-8B} is fine-tuned on 1,228 SFT training instances derived from PersonaArena trajectories. Specifically, trajectories generated by different models are ranked by their overall evaluation scores, and the top 50 complete trajectories are retained. These trajectories are then decomposed into behavior-level instances, each consisting of a prompt and the corresponding protagonist response at a single step. After conversion into the supervised fine-tuning format, we obtain 1,228 training instances for optimization. As shown in Fig.~\ref{fig:fine-tune}, the resulting \textbf{SFT-Qwen3-8B} consistently improves over the base Qwen3-8B across all eight evaluation dimensions, with an average gain of approximately 21.96\%.  In particular, it shows substantial improvements in IR, BA, and BC, with
  gains of 32.07\%, 30.17\%, and 27.86\%, respectively. Notably, it outperforms {GPT-4.1} in BA, IM, and BC. These results suggest that supervised imitation of high-quality trajectories effectively enhances the richness of character
  interactions and behavioral consistency.

\textbf{DPO fine-tuning.} Building on the SFT-initialized Qwen3-8B, we further perform preference optimization using 665 DPO training pairs. Specifically, we construct candidate trajectory pairs from complete trajectories generated by different models for the same persona, rank them by the gap in their overall evaluation scores, and retain the top 50 pairs. Each selected pair is then decomposed into aligned behavior-level instances under the same prompt, where the higher-quality response is treated as the \emph{chosen} response and the lower-quality one as the \emph{rejected} response. Converting these comparisons into the standard DPO format yields 665 preference training pairs. As illustrated in Fig.\ref{fig:fine-tune}, the resulting \textbf{DPO-Qwen3-8B} achieves the best overall performance, with an average gain of approximately 27.83\% over the base model. In particular, it shows remarkable improvements in IR, BA, and AD. Compared to SFT-Qwen3-8B, DPO further improves by 5.21\%, with notable gains in IR (15.71\%) and AD (14.67\%). Notably, it surpasses {GPT-4.1} in overall performance, outperforming it in six dimensions including KA, BA, AD, BC, EE, and IR. This suggests that preference-based trajectory optimization helps the model better capture implicit behavioral preferences and produce more contextually appropriate, emotionally nuanced, and human-like role-playing behaviors.

Overall, these findings confirm that learning from high-quality trajectories significantly enhances the role-playing capability of large language models. The SFT paradigm provides explicit behavioral demonstrations for imitation, while the DPO paradigm further refines model alignment with implicit human preferences. Together, they validate the effectiveness of trajectory-based fine-tuning in fostering contextually coherent, emotionally expressive, and personality-consistent character behaviors.
Cross-framework transfer results on external benchmarks are reported in Appendix~\ref{sec:app_external_eval}.

\subsection{Analysis of Evaluation Stages}

\paragraph{Impact of NPC Model Capability.}

To examine the robustness of our evaluation framework, we analyze whether the capability of non-player character (NPC) models affects the measured role-playing performance. In the default configuration, all NPCs and the \textit{Environment Agent} are instantiated using \textit{Qwen3-32B}, which provides stable contextual reasoning and balanced dialogue behavior. To test the sensitivity of our framework, we replace all NPCs with a stronger closed-source model, \textit{GPT-4.1}, while keeping the protagonist model, evaluation protocol, and scenario settings identical.

\begin{figure}[t]
\centering
\includegraphics[width=\linewidth]{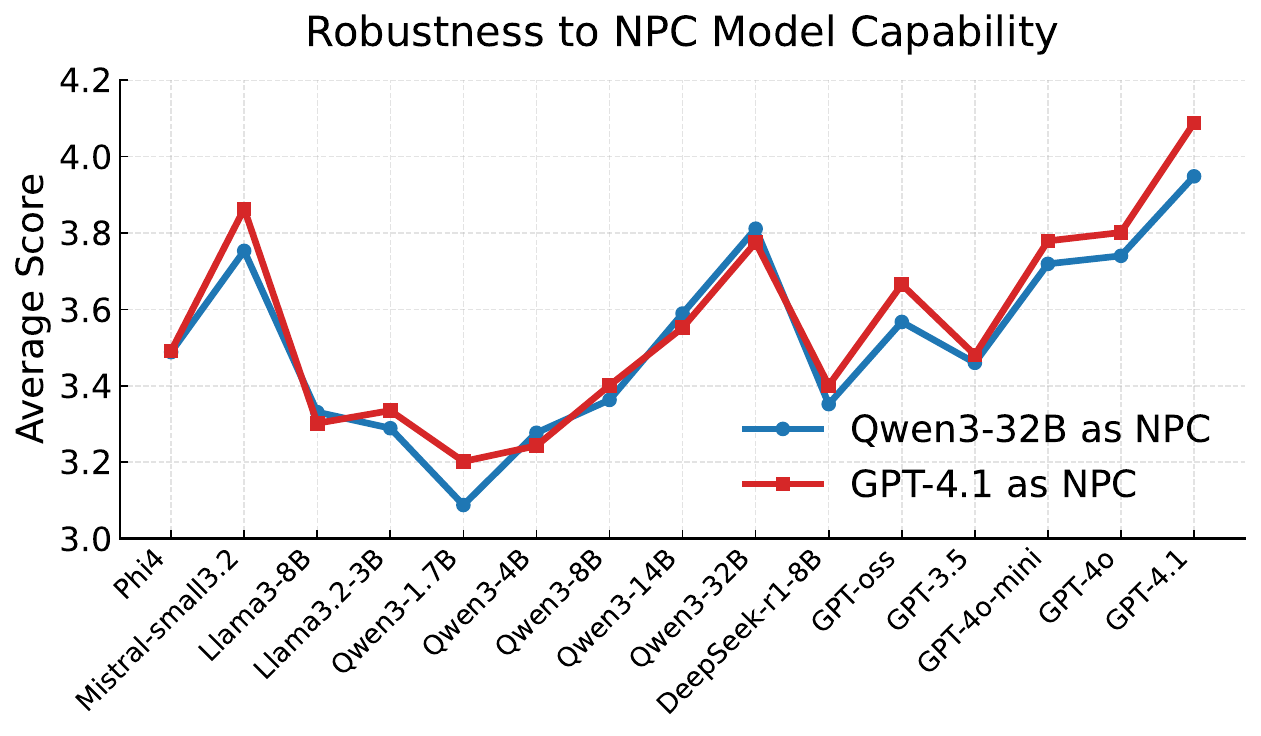}
\caption{Performance Stability under Different NPC Configurations}
\label{fig:npc-sensitivity}
\end{figure}

As shown in Fig.~\ref{fig:npc-sensitivity}, the overall performance curves under both configurations remain highly consistent, with only minor deviations across scenes. This stability demonstrates that the proposed multi-judge debate evaluation is largely unaffected by changes in NPC capability. While GPT-4.1 as NPC yields slightly higher overall scores, the relative rankings and performance patterns across scenarios remain stable. These findings confirm that our framework reliably isolates protagonist performance from NPC variability, ensuring fair and reproducible evaluation even when interactive agents differ in capability.
Additional robustness ablations are reported in Appendix~\ref{sec:app_ablation}.

\section{Related Work}

\subsection{Character-Level Role-Playing}
Character-level role-playing explores how LLMs impersonate specific fictional, historical, or celebrity figures with coherent personalities and linguistic styles. Early work such as ChatHaruhi~\cite{li2023chatharuhi} revive anime characters through curated dialogues, while Character-LLM~\cite{shao-etal-2023-character} and CharacterGLM~\cite{zhou2023characterglm} advance model customization via fine-tuning and conditioning.
Recent benchmarks broaden evaluation scope and granularity. RoleLLM~\cite{wang-etal-2024-rolellm} presents \textsc{RoleBench} with 100 profiles and 168K dialogues, introducing Role-Conditioned Instruction Tuning for improved role fidelity. CharacterEval~\cite{tu2024charactereval} contributes a multi-turn Chinese dataset with 77 literary figures and learned metrics for persona consistency. InCharacter~\cite{wang-etal-2024-incharacter} assesses psychological fidelity via interview-style Q\&A, showing linguistic imitation alone is insufficient. SocialBench~\cite{chen2024socialbench} evaluates sociality across 500 characters and 6K prompts, measuring empathy and cooperation.
Interactive environments now capture dynamic behaviors. CharacterBox~\cite{wang-etal-2025-characterbox} offers a text-based world for evaluating character traits through simulated interactions, and Xu et al.~\cite{xu2024character} study persona-driven moral reasoning. The large-scale CoSER~\cite{wang2025coser} dataset aggregates 18K literary characters and trains 8B/70B models surpassing prior baselines. Other earlier datasets~\cite{chen2022large,shen2023roleeval,li2023chatharuhi} also provide valuable foundations. However, fictional characters often display exaggerated traits unlike real people. Our work instead targets broader \emph{persona-level} roles that capture realistic social dynamics.

\subsection{Persona-Level Role-Playing}
Persona-based role-playing focuses on general social archetypes rather than fixed identities, emphasizing consistency with enduring traits such as occupation and values. A recent survey~\cite{chen2024persona} reviews progress in persona-grounded dialogue and evaluation. Persona-Chat~\cite{zhang-etal-2018-personalizing} lay the foundation for this field, later extended by \citet{jandaghi2023faithful} with Synthetic-Persona-Chat, an unsupervised dataset featuring 5K personas and 20K dialogues.
Persona Hub~\cite{ge2024scaling} introduces one billion automatically generated personas for data synthesis, and OpenCharacter~\cite{wang2025opencharacter} leverages them to train customizable role-playing LLMs. RoleCraft-GLM~\cite{tao2023rolecraft} enriches personas with detailed emotional attributes, while DMT-RoleBench~\cite{yuan2025dmt} benchmarks dynamic, intent-driven dialogue performance. \cite{peng2024quantifying} focus on evaluating the faithfulness (i.e., whether LLMs violate the constraints of the persona) in role-play. \cite{samuel2024personagym} use dynamic QAs with the decision theory for assessing persona agents.

Persona-level role-playing advances enable large-scale social simulations and behavioral studies~\cite{aher2023using,hewitt2024predicting}. Toolkits such as TinyTroupe~\cite{salem2025tinytroupe} and AgentSociety~\cite{piao2025agentsociety} deploy persona-grounded agents to model emergent social behavior. Yet, reliable evaluation of persona-based role-playing in realistic, interactive contexts remains underexplored. Our work addresses this gap by assessing LLMs’ social role-playing competence within dynamic simulated environments.

\section{Conclusions}
In this work, we presented PersonaArena, a dynamic simulation framework for evaluating and improving persona-level role-playing in LLMs. By constructing a nuanced persona bank from user-generated social content and eliciting multi-turn, context-rich interactions, PersonaArena enables rigorous assessment of LLMs’ fidelity, coherence, and adaptability through a multi-agent judge. Our experiments show that PersonaArena not only facilitates robust evaluation but also provides valuable post-training data, leading to enhanced role-playing consistency and realism. We believe PersonaArena lays the groundwork for advancing more authentic and socially adept AI agents, and encourages research in dynamic, context-driven evaluation and training methodologies.

\section{Limitations}
Although our multi-judge debate framework helps mitigate the biases associated with individual automated judges, it does not fully achieve the accuracy or nuance of ideal human judgment. The aggregation of LLM-based opinions can still reflect underlying model biases, and subtle aspects of persona fidelity may be missed or misinterpreted. Future work could incorporate more diverse judge models, hybrid human-AI evaluation, or improved aggregation methods to further narrow this gap.

Our study primarily addresses the technical aspects of role-playing faithfulness and consistency without considering the ethical implications of certain roles, such as those associated with harmful, antisocial, or “evil” behavior. Whether LLMs should be capable of convincingly simulating such roles remains an open and complex question, involving broader societal and safety considerations. In this work, we focus on role-playing capability itself and leave the exploration of normative boundaries and safeguards for future research.

The selection and configuration of NPC LLMs used to drive interactions within PersonaArena can influence the quality and diversity of elicited behaviors. Differences in model architecture, training data, or prompt design may affect NPC responses and, consequently, the evaluation of the protagonist LLM. While we aim to use the strongest available models from both closed-source and open-source sources, the impact of NPC selection on simulation outcomes requires further systematic study.

Our persona bank is constructed from user-generated social content, which may be subject to demographic, cultural, or platform-specific biases. These biases could limit the representativeness and generalizability of the personas and interactions modeled in PersonaArena. Expanding the dataset to include more diverse sources and conducting bias audits are important directions to enhance fairness and inclusivity.

\section*{Acknowledgments}
This work was motivated by the Society Zero Universe platform's need for agents with robust cognitive consistency and high fidelity in role-playing.

This work is supported by the National Natural Science Foundation of China (Grant No. 62276171, 62476173, 62532007), the Guangdong Basic and Applied Basic Research Foundation (Grant No. 2024A1515011938 and 2020B1515120028), the Shenzhen Fundamental Research Project (Grant No. ZDCY20250901110940006, JCYJ20-240813141503005, JCYJ20240813142610014), and the Major Special Project for Philosophy and Social Sciences Research of the Ministry of Education (Grant No. 2025JZDZ010).

% Bibliography entries for the entire Anthology, followed by custom entries
%\bibliography{anthology,custom}
% Custom bibliography entries only
\bibliography{custom}

\clearpage
\onecolumn
\appendix
\section{Evaluation Information}

\subsection{Evaluation Metrics}
\label{sec:app_metrics}

The full metrics in evaluation engine are organized into four categories:

\begin{itemize}[leftmargin=*]

\item\textbf{Character Fidelity.}  
This dimension evaluates how faithfully the model reproduces the character's knowledge and behavioral patterns.  

\textit{Knowledge Accuracy (KA):} measures factual correctness consistent with the character’s background

\textit{Behavioral Accuracy (BA):} assesses the alignment between the model’s actions and the predefined personality traits.

\item \textbf{Emotional Expressiveness.}  
This category examines the human-likeness and emotional engagement of the role-play.  

\textit{Emotional Expression (EE):} evaluates the vividness and appropriateness of affective communication

\textit{Personality Traits (PT):} measures the model's ability to maintain stable personality features across interactions.

\item \textbf{Interactive Coherence.}  
This dimension captures the logical and temporal continuity of behavior throughout the conversation. 

\textit{Immersion (IM):} quantifies the model’s ability to remain in character, maintaining narrative consistency.

\textit{Behavioral Coherence (BC):} measures causal and contextual alignment between past and ongoing actions.

\item \textbf{Behavioral Diversity.}  
This category assesses the diversity and richness of the model’s social expressions and goal-oriented behaviors. 

\textit{Adaptability (AD):} evaluates how flexibly the model adjusts to evolving contexts while preserving persona consistency.

\textit{Interaction Richness (IR):} measures the breadth and depth of its social responses, reflecting creative and varied role-play performance.

\end{itemize}

\begin{figure}[!ht]
    \centering
    \includegraphics[width=\linewidth]{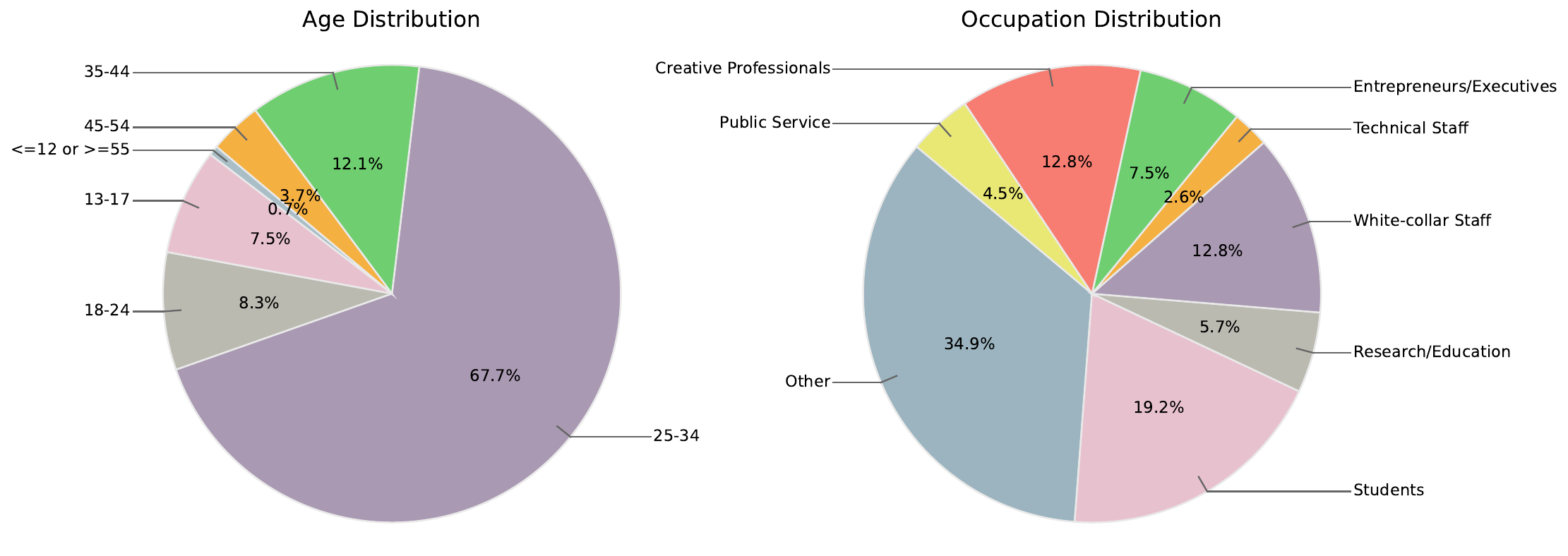}
    \caption{Distributions of age groups and occupation categories in Persona Bank.}
    \label{fig:PB_info}
\end{figure}

\section{Persona Bank Information}
\label{sec:app_PBI}

The persona bank comprises 1,000 characters derived from real social media data, spanning six age groups and eight occupational categories. The overall statistics are presented in Figure~\ref{fig:PB_info}.

\paragraph{Demographic summary.}
For the 1,000-persona subset, the age distribution is: $\leq12$ or $\geq55$ (0.7\%), 13--17 (7.5\%), 18--24 (8.3\%), 25--34 (67.7\%), 35--44 (12.1\%), and 45--54 (3.7\%). Occupation distribution includes Students (19.2\%), White-collar Staff (12.8\%),
  Creative Professionals (12.8\%), Entrepreneurs/Executives (7.5\%), Research/Education (5.7\%),
  Public Service (4.5\%), Technical Staff (2.6\%), and Other (34.9\%).

The detailed attributes of a sample persona are as follows:

\begin{figure}[!h]
    \centering
    \includegraphics[width=\linewidth]{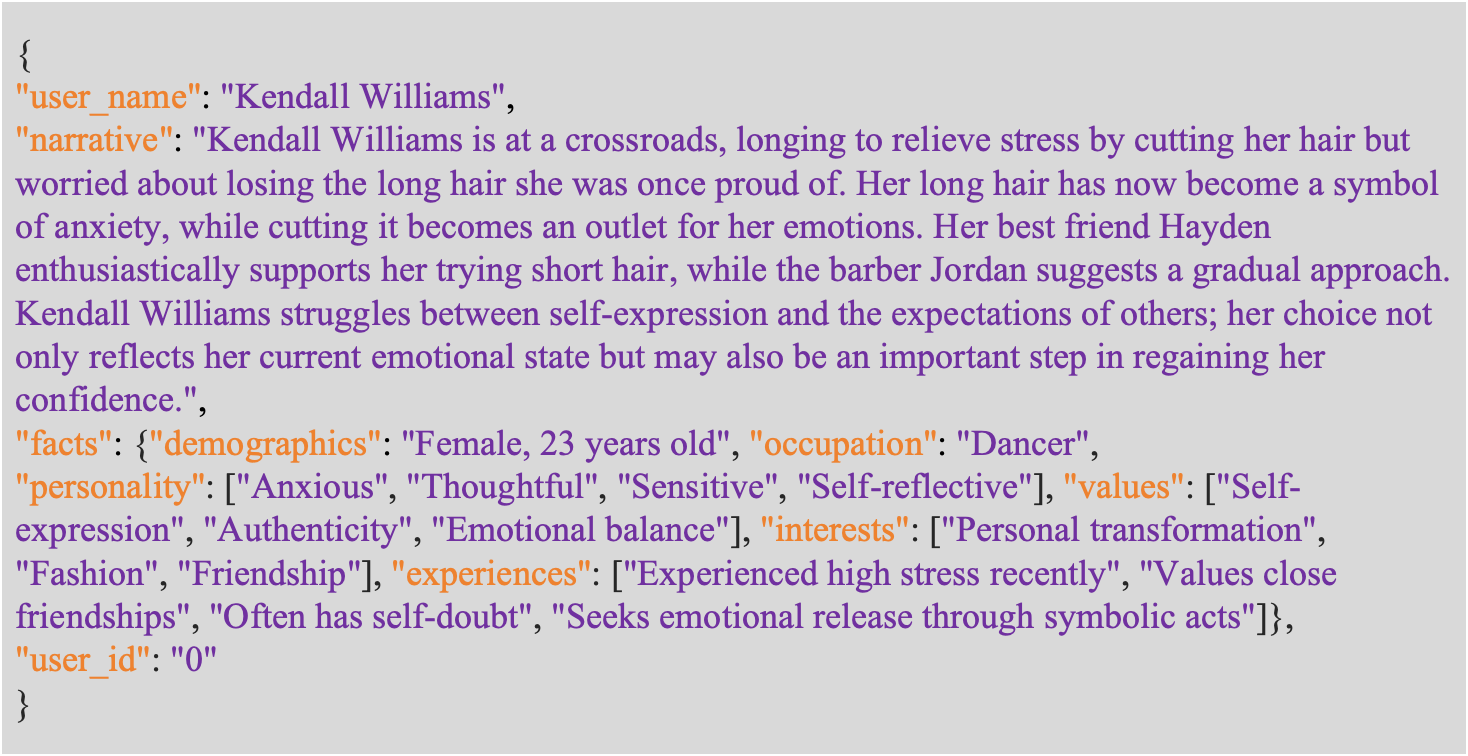}
    \caption{Example persona card with narrative and structured facts.}
    \label{fig:persona_info}
\end{figure}

\paragraph{Data construction and anonymization quality.}
To complement the persona statistics, we provide additional quality-control evidence from anonymization analyses.

\begin{table}[!ht]
\centering
\small
\caption{Human discrimination between original and anonymized persona cards (20 pairs, 10 annotators).}
\label{tab:app_anon_human}
\begin{tabular}{lcc}
\toprule
\textbf{Choice Type} & \textbf{Count} & \textbf{Proportion} \\
\midrule
Original persona chosen   & 94  & 47.0\% \\
Anonymized persona chosen & 106 & 53.0\% \\
\bottomrule
\end{tabular}
\end{table}

In this task, each original--anonymized pair was presented in random order, and 10 annotators selected which card better matched the underlying author profile. The near-balanced preference (47.0\% vs.\ 53.0\%) indicates that anonymization preserves persona semantics while obscuring direct identity cues.

\begin{table}[!ht]
\centering
\small
\caption{Effect of anonymization on average model scores.}
\label{tab:app_anon_scores}
\begin{tabular}{lccc}
\toprule
\textbf{Model} & \textbf{Post-anonymization} & \textbf{Pre-anonymization} & \textbf{Diff (Post - Pre)} \\
\midrule
Phi-4             & 3.769 & 3.725 & 0.044 \\
Mistral-small-3.2 & 3.772 & 3.817 & -0.045 \\
Qwen3-32B         & 3.982 & 3.901 & 0.081 \\
GPT-4.1           & 4.018 & 4.046 & -0.028 \\
\bottomrule
\end{tabular}
\end{table}

We further ran the full evaluation pipeline on 50 randomly sampled original--anonymized persona pairs across four model families. Score shifts are small (about 0.02--0.08 absolute points), and the relative ranking is unchanged (GPT-4.1 $>$ Qwen3-32B $>$ Mistral-small-3.2/Phi-4), indicating limited impact of anonymization on comparative conclusions.

\section{Social Scene Information}
\label{sec:app_SCI}

The simulated scenes in CharacterArena are generated by the scene generator based on persona information. The social scenarios included in the experiments fall into the following categories:

\begin{table}[!ht]
\centering
\caption{Scenario categories and representative examples.}
\begin{tabular}{p{4cm}p{10cm}}
\hline
\textbf{Scenario Category} & \textbf{Representative Examples} \\
\hline
Public Settings & Taking public transport; attending a concert; visiting a museum. \\
Family Scenarios & Dining with family; caring for a child; resolving household issues. \\
Education \& Learning & Classroom discussion; group project collaboration; online lecture participation. \\
Cross-cultural \& Interdisciplinary Contexts & International academic exchange; bilingual communication; cross-domain innovation workshop. \\
Friends \& Family Gatherings & Birthday celebration; reunion dinner; casual weekend outing. \\
Community \& Neighborhood & Volunteering in a local event; neighborhood meeting; helping neighbors. \\
Online Social Interaction & Chatting on social media; participating in online forums; video conferencing. \\
Ceremonies \& Public Relations & Wedding ceremony; press conference; public speech or award event. \\
Formal Business Settings & Job interview; client negotiation; professional presentation. \\
Workplace Daily Life & Team meeting; coworker collaboration; dealing with work stress. \\
\hline
\end{tabular}
\label{tab:scenario_examples}
\end{table}

The specific information of a generated social scene is as follows:

\begin{figure}[!htbp]
    \centering
    \includegraphics[width=\linewidth]{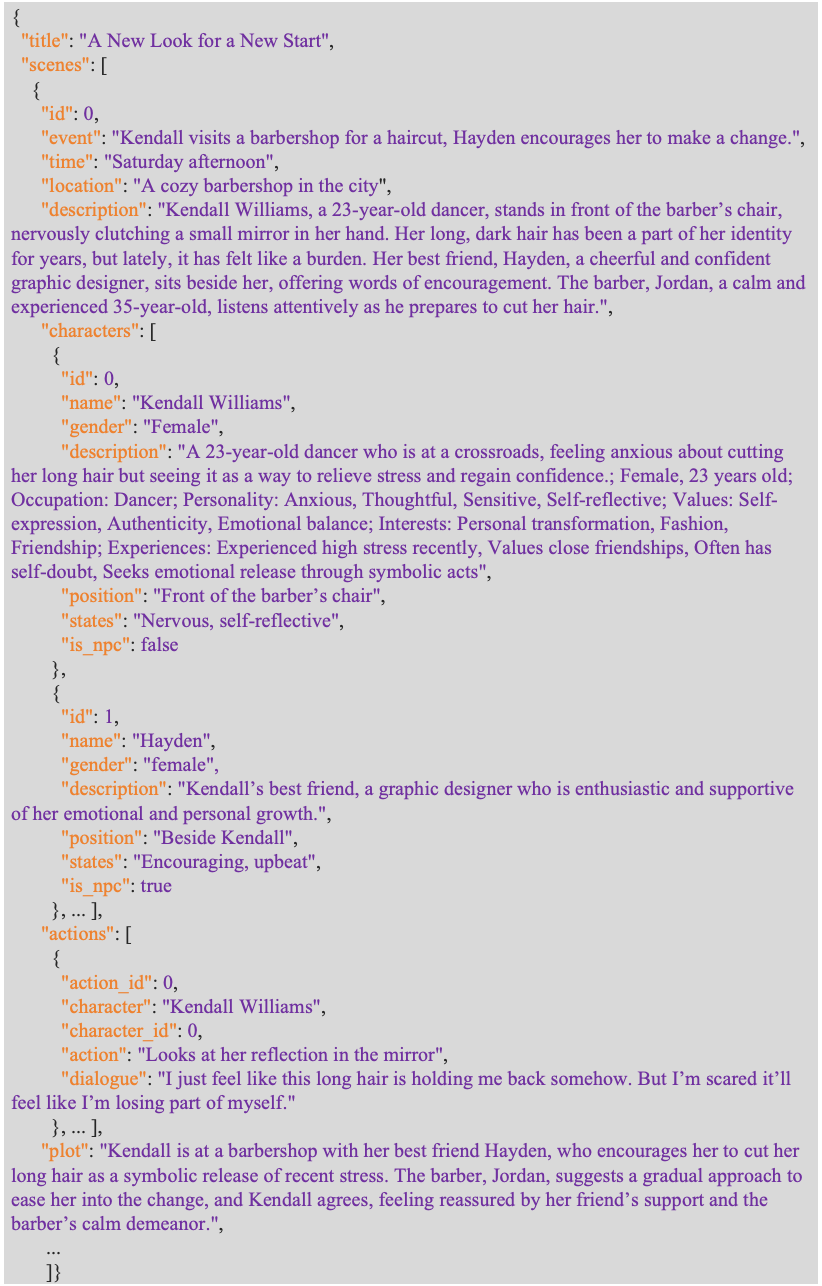}
    \caption{Example generated social-scene specification used for simulation initialization.}
    \label{fig:scene_info}
\end{figure}

\clearpage

\section{Detailed Prompt}

\label{sec:app:prompt}

\subsection{Persona Agent Prompt}
\textbf{Action:} Design environment-appropriate actions for each character based on their personality traits and perception of the current scene, thereby fostering realistic social simulations. 
\begin{figure}[!ht]
    \centering
    \includegraphics[width=\linewidth]{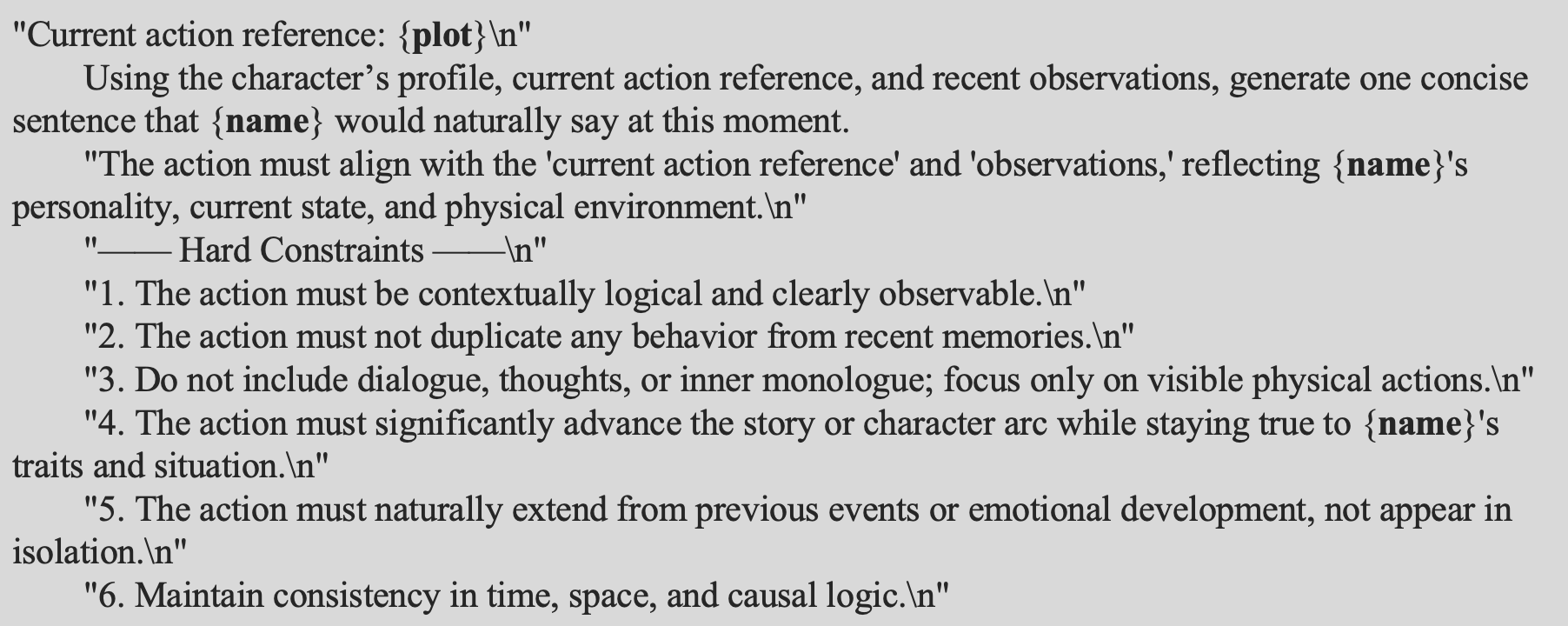}
    \caption{Persona Agent prompt for action generation.}
    \label{fig:app_prompt_action}
\end{figure}

\textbf{Dialogue:} Generate dialogues grounded in the character’s personality, environmental perception, interactions with NPCs, and recent memories.
\begin{figure}[!ht]
    \centering
    \includegraphics[width=\linewidth]{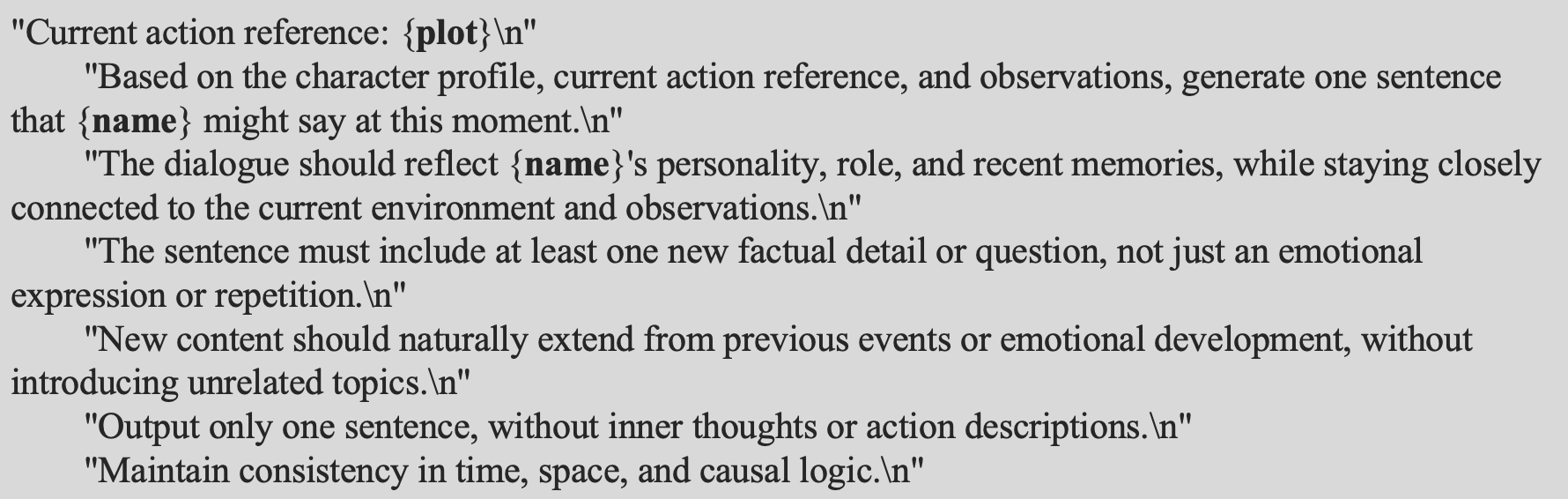}
    \caption{Persona Agent prompt for dialogue generation.}
    \label{fig:app_prompt_dialogue}
\end{figure}

\textbf{Reaction:} Describe the character’s reaction—whether through actions or dialogue—based on their observations. characters.
\begin{figure}[!ht]
    \centering
    \includegraphics[width=\linewidth]{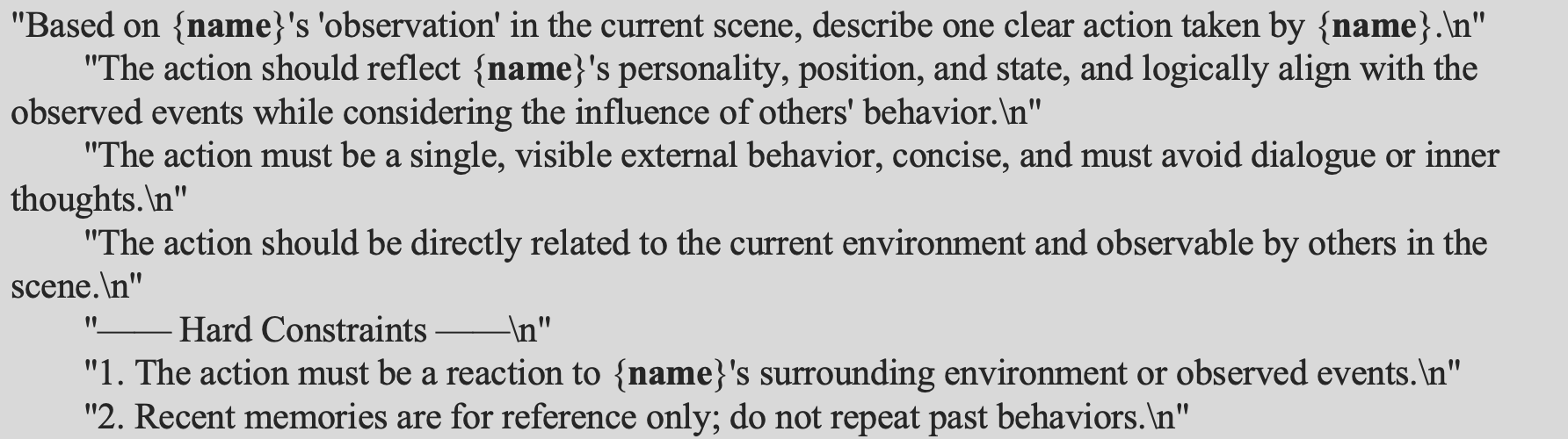}
    \caption{Persona Agent prompt for reaction generation.}
    \label{fig:app_prompt_reaction}
\end{figure}

\clearpage

\subsection{Environment Agent Prompt}

\textbf{Scenario Setup:} Generate concrete social scenes based on persona records, incorporating the basic information of characters and NPCs, along with their corresponding initial actions or dialogues. 

\begin{figure}[!ht]
    \centering
    \includegraphics[width=\linewidth]{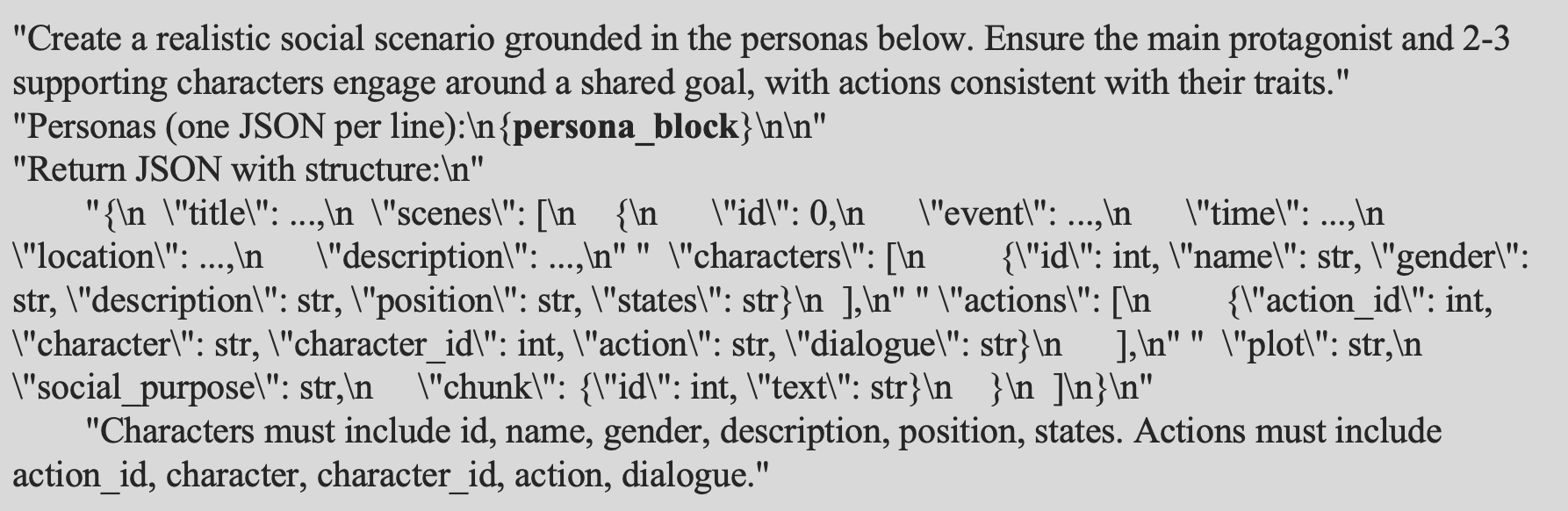}
    \caption{Environment Agent prompt for scenario setup.}
    \label{fig:app_prompt_scene_setup}
\end{figure}

\textbf{Analyze Influence:} Analyze and describe the practical impact that one character's current physical actions or words might have on another character. 

\begin{figure}[!ht]
    \centering
    \includegraphics[width=\linewidth]{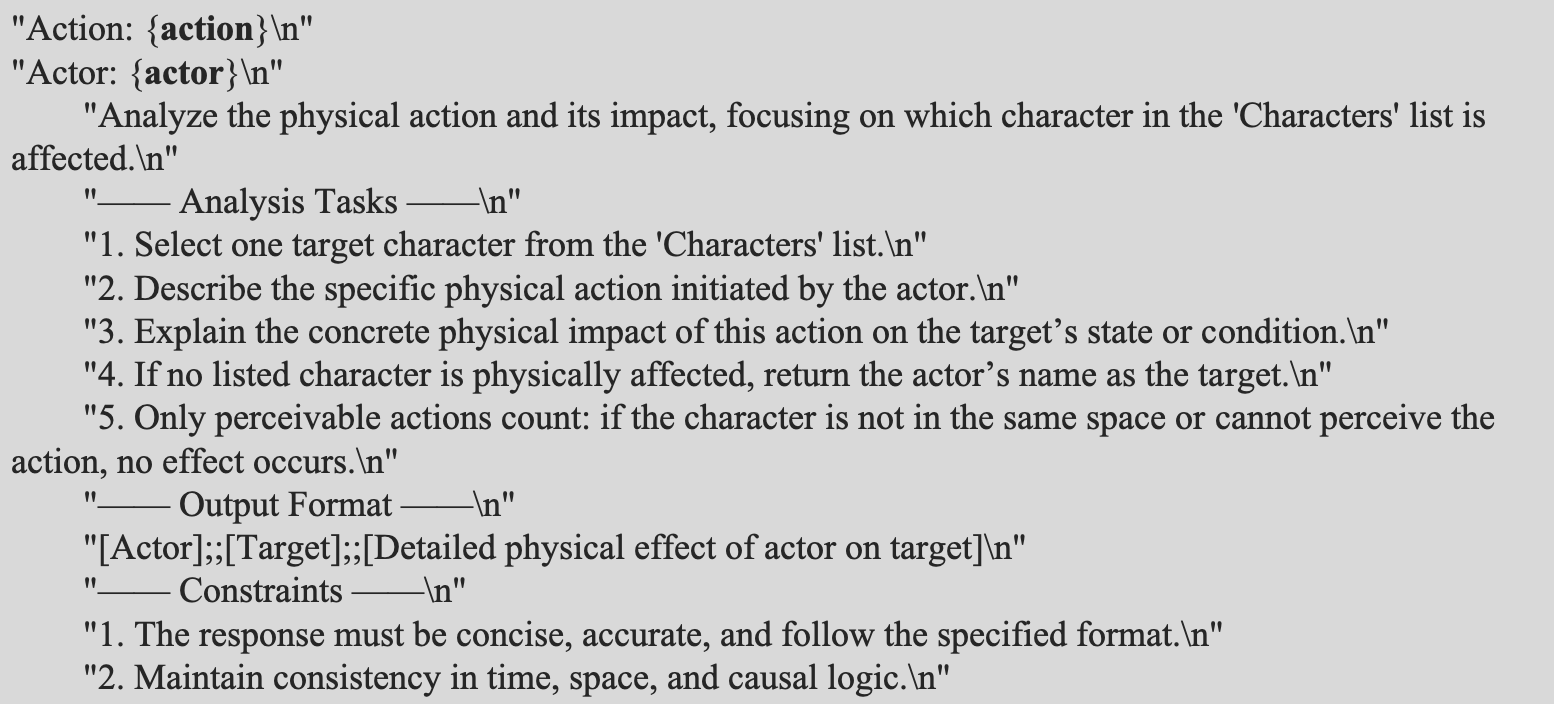}
    \caption{Environment Agent prompt for influence analysis.}
    \label{fig:app_prompt_influence}
\end{figure}

\clearpage
\onecolumn
\textbf{Adaptive Turn Control:} By checking whether the protagonist has reached the preset checkpoints in the current interaction, we can determine if the next round should continue or end. 

\begin{figure}[!ht]
    \centering
    \includegraphics[width=0.8\linewidth]{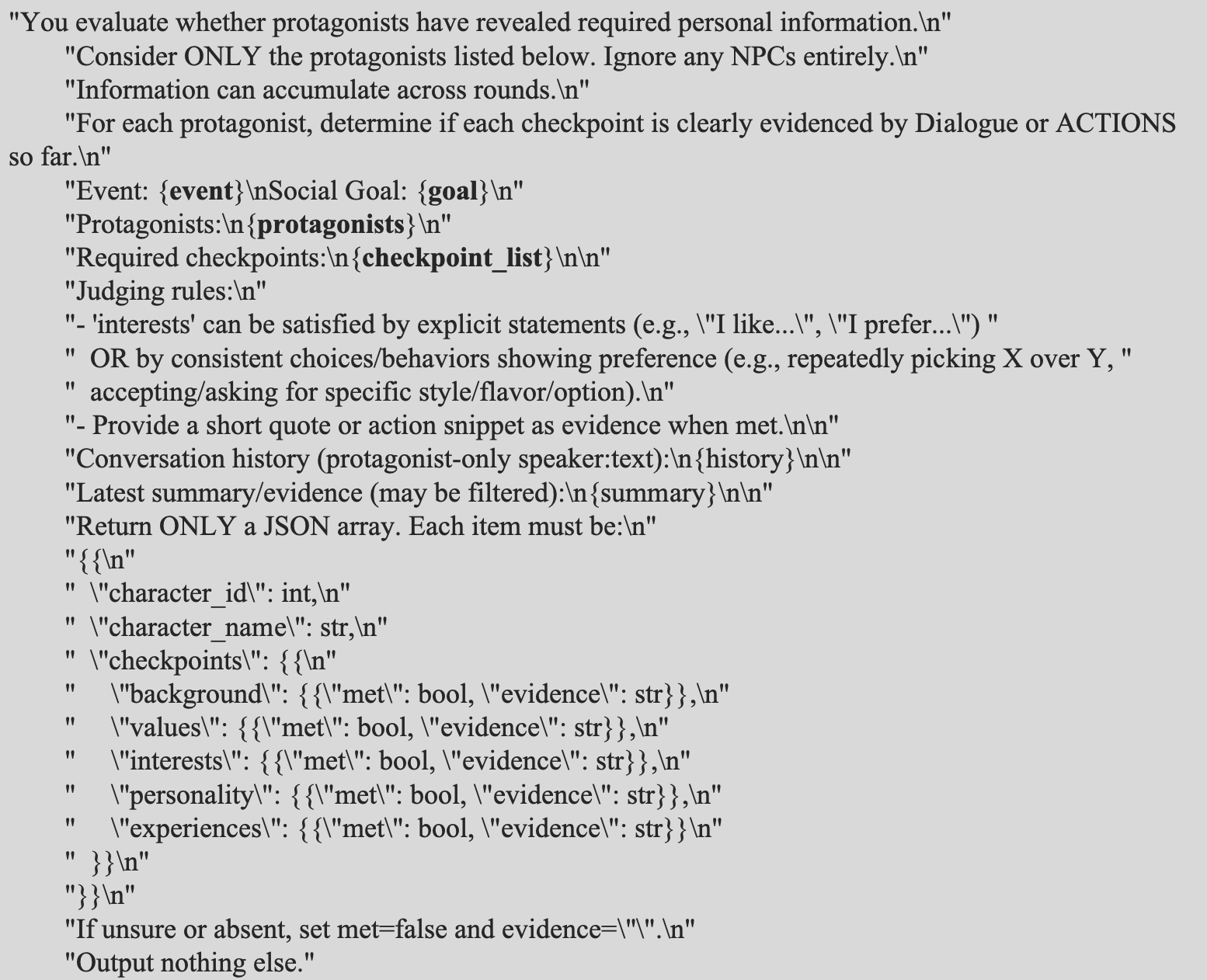}
    \caption{Environment Agent prompt for adaptive turn control.}
    \label{fig:app_prompt_turn_control}
\end{figure}

\textbf{Character State Update:} By analyzing the scene context and interaction progress from the previous round, the character’s current position, state, and environmental observations are updated. 

\begin{figure}[!ht]
    \centering
    \includegraphics[width=\linewidth]{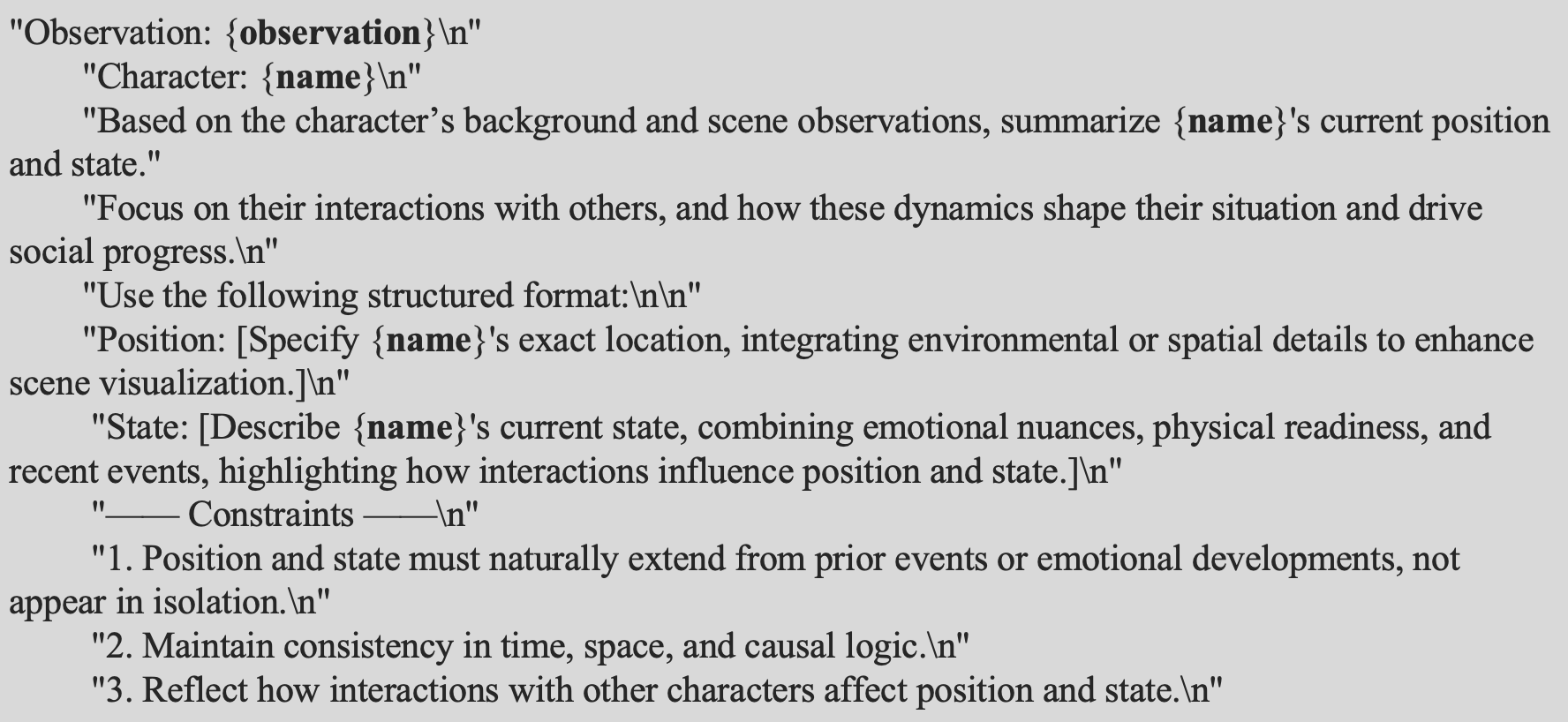}
    \caption{Environment Agent prompt for character-state update.}
    \label{fig:app_prompt_character_update}
\end{figure}

\textbf{Environment Update:} Make adjustments to the physical environment based on observations from the previous round of interactions. 

\begin{figure}[!ht]
    \centering
    \includegraphics[width=\linewidth]{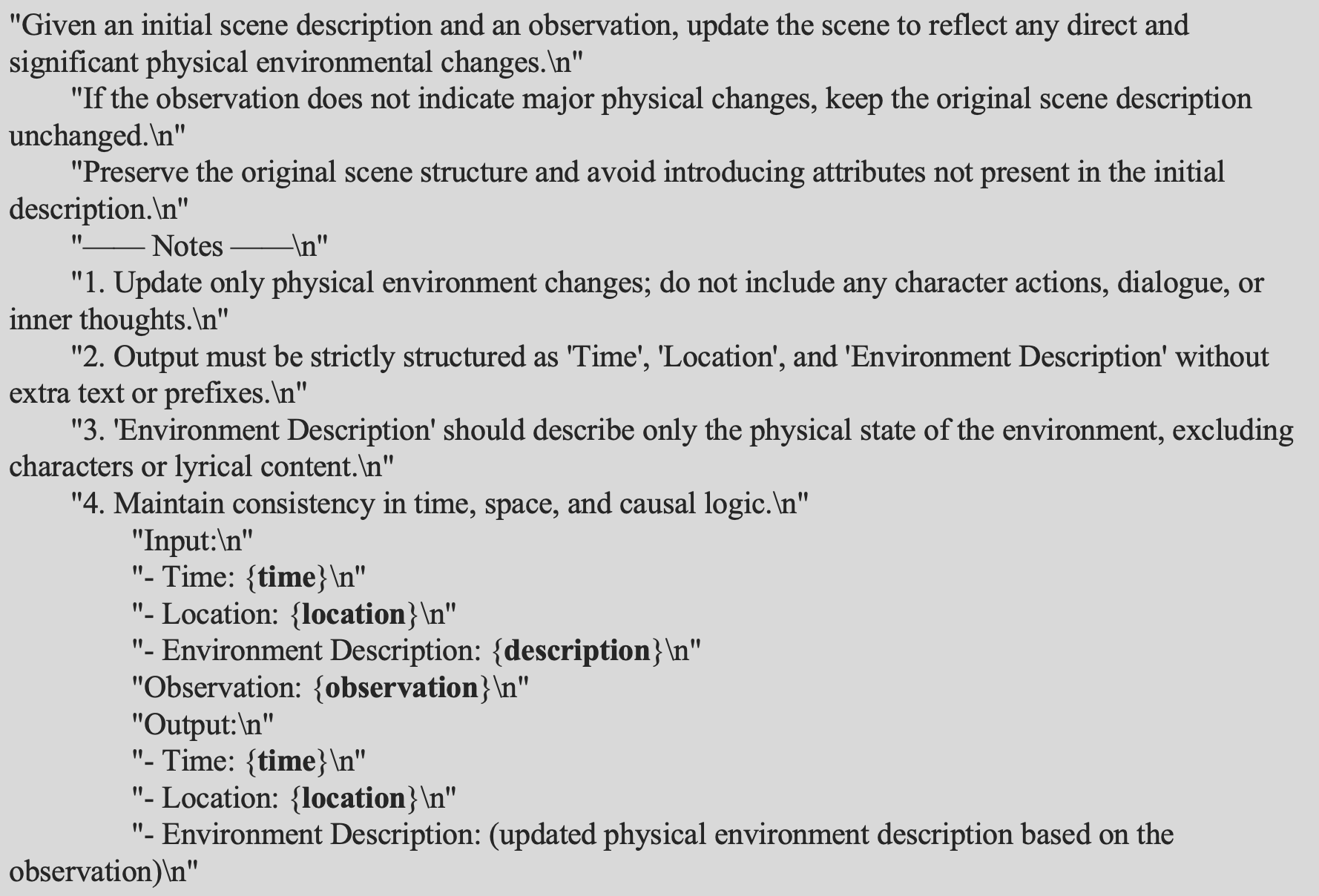}
    \caption{Environment Agent prompt for environment update.}
    \label{fig:app_prompt_env_update}
\end{figure}

% \section{Experimental Details}

% \section{Experiment Compute Resources}

\section{Human Evaluation and Judge Calibration Details}
\label{sec:app_human_eval}

\subsection{Human evaluation}
\label{sec:app_he}

Human evaluation uses 4 graduate annotators with prior role-playing/simulation experience. Each trajectory is rated on KA, BA, EE, PT, IM, BC, AD, and IR with a 1--5 Likert scale. Every trajectory is independently rated by at least three annotators, and final scores are averaged. Before formal annotation, annotators jointly review guidelines and pilot examples for calibration. 

\subsection{Multi-Judge Calibration Case Study}
\label{sec:app_judge_case}

\begin{table}[!ht]
\centering
\small
\caption{Average scores from six judges for four evaluated models (3 representative trajectories per model).}
\label{tab:app_judge_case}
\begin{tabular}{lcccccc}
\toprule
\textbf{Model} & \textbf{DeepSeek-R1} & \textbf{Mistral-3.2} & \textbf{Llama-3.1-8B} & \textbf{Qwen3-32B} & \textbf{GPT-4o} & \textbf{Phi-4} \\
\midrule
Phi-4        & 4.50 & 3.92 & 4.00 & 3.33 & 3.29 & 3.96 \\
Mistral-small-3.2     & 4.50 & 3.92 & 4.00 & 3.46 & 3.50 & 3.98 \\
Qwen3-32B & 4.50 & 4.25 & 4.33 & 3.92 & 3.88 & 4.08 \\
GPT-4.1             & 4.46 & 4.38 & 4.13 & 3.88 & 4.17 & 3.96 \\
\bottomrule
\end{tabular}
\end{table}

This case study shows clear that different judges have very different absolute calibration (e.g., DeepSeek tends to be consistently more lenient, whereas Qwen3-32B and GPT-4o are more conservative), confirming that any single LLM judge comes with its own scoring scale and stylistic bias. At the same time, when looking across models, the relative quality ordering is quite stable: trajectories that are strong for one judge are also strong for the others, and no model’s ranking flips dramatically under a different judge.

This is precisely where the multi-judge aggregation is useful: it does not overturn the underlying consensus about model quality, but smooths out individual leniency/harshness and model-family-specific preferences, producing a more robust and human-aligned signal than relying on a single judge such as Qwen3-32B. This is particularly important when these scores are later used as training signals for SFT/DPO, where overfitting to one model’s quirks would directly harm generalization.

\section{Cross-framework Generalization}
\label{sec:app_external_eval}

To assess cross-framework generalization beyond PersonaArena, we additionally evaluate the base and fine-tuned Qwen3-8B models on two external persona/role-playing benchmarks: PersonaGym and the RoleBench suite from RoleLLM. In both settings, we instantiate agents using personas extracted from PersonaArena as underlying profiles, while strictly following each benchmark’s official evaluation protocol (PersonaScore over five decision-theoretic tasks in PersonaGym, and GPT-4–based win rate in RoleBench):

\begin{table}[!ht]
\centering
\small
\caption{External evaluation on PersonaGym.}
\label{tab:app_personagym}
\begin{tabular}{lcccccc}
\toprule
\textbf{Model} & \textbf{Action Just.} & \textbf{Expected Action} & \textbf{Ling. Habits} & \textbf{Persona Cons.} & \textbf{Toxicity Ctrl.} & \textbf{PersonaScore} \\
\midrule
Qwen3-8B      & 3.38 & 3.13 & 3.13 & 3.75 & 4.91 & 3.66 \\
SFT-Qwen3-8B  & 3.50 & 3.63 & 3.50 & 3.88 & 4.93 & 3.88 \\
DPO-Qwen3-8B  & 3.88 & 3.63 & 3.75 & 4.25 & 4.92 & 4.09 \\
GPT-4.1       & 4.13 & 4.13 & 4.00 & 4.25 & 4.88 & 4.28 \\
\bottomrule
\end{tabular}
\end{table}

\begin{table}[!ht]
\centering
\small
\caption{External evaluation on RoleBench (GPT-4-based win rate).}
\label{tab:app_rolebench}
\begin{tabular}{lc}
\toprule
\textbf{Model} & \textbf{Win Rate (\%)} \\
\midrule
Qwen3-8B      & 0.0 \\
SFT-Qwen3-8B  & 28.6 \\
DPO-Qwen3-8B  & 37.1 \\
GPT-4.1       & 34.3 \\
\bottomrule
\end{tabular}
\end{table}

On PersonaGym, both SFT-Qwen3-8B and DPO-Qwen3-8B consistently outperform the base Qwen3-8B across all reported dimensions (Action Justification, Expected Action, Linguistic Habits, Persona Consistency, Toxicity Control). The overall PersonaScore increases from 3.66 (base) to 3.88 (SFT) and 4.09 (DPO), further narrowing the gap to GPT-4.1 (4.28). On RoleBench, with GPT-4o as the evaluator, the win rate of Qwen3-8B improves from 0.0\% (base) to 28.6\% (SFT) and 37.1\% (DPO), with DPO-Qwen3-8B even slightly surpassing GPT-4.1 (34.3\%).

These results indicate that the gains from our SFT/DPO training are not confined to the PersonaArena pipeline: they transfer to independently designed benchmarks as more persona-faithful behavior and style imitation.

\section{Additional Robustness Experiments}
\label{sec:app_ablation}

\subsection{Arbiter Sensitivity}

We evaluate whether changing the arbiter model affects conclusions. All settings are fixed except the arbiter choice (same personas, prompts, judge set, and evaluation pipeline).

\begin{table}[!ht]
\centering
\small
\caption{Average scores under different arbiters.}
\label{tab:app_arbiter}
\begin{tabular}{lccccc}
\toprule
\textbf{Model} & \textbf{GPT-4o} & \textbf{Qwen3-32B} & \textbf{Qwen3-1.7B} & \textbf{Diff (4o-32B)} & \textbf{Diff (4o-1.7B)} \\
\midrule
Phi-4             & 3.781 & 3.798 & 4.072 & -0.017 & -0.291 \\
Mistral-small-3.2 & 3.875 & 3.814 & 4.120 & 0.061  & -0.245 \\
Qwen3-32B         & 3.887 & 3.844 & 3.951 & 0.043  & -0.064 \\
GPT-4.1           & 4.004 & 4.025 & 4.262 & -0.021 & -0.258 \\
\bottomrule
\end{tabular}
\end{table}

Overall, these results suggest that our conclusions are robust to the choice of arbiter. Swapping GPT-4o for a similarly strong arbiter (Qwen3-32B) leads to only very small changes in absolute scores and leaves the qualitative picture unchanged: GPT-4.1 remains the top-performing model, and the relative gaps between models are broadly similar.

Using a much weaker arbiter (Qwen3-1.7B) tends to inflate all scores and blur some of the distinctions between models, likely because its limited capability fails to reliably catch more subtle errors. Thus, the arbiter mainly affects the scale of the scores rather than the ranking or the overall conclusions drawn from the benchmark.

\subsection{Inter-Judge Composition Sensitivity}

We replace the original judge set with two alternatives while keeping all other settings unchanged. Set A uses \{GPT-4o, Mistral-small-3.2, Phi-4\}, and Set B uses \{Llama-3.1-8B, Phi-4, Qwen3-32B\}, to test cross-family judge composition effects.

\begin{table}[!ht]
\centering
\small
\caption{Average scores under different judge-set compositions.}
\label{tab:app_judgeset}
\begin{tabular}{lccccc}
\toprule
\textbf{Model} & \textbf{Original} & \textbf{Set A} & \textbf{Set B} & \textbf{Diff (A-Orig)} & \textbf{Diff (B-Orig)} \\
\midrule
Phi-4             & 3.781 & 3.801 & 3.958 & 0.020 & 0.177 \\
Mistral-small-3.2 & 3.875 & 3.875 & 4.042 & 0.000 & 0.167 \\
Qwen3-32B         & 3.887 & 3.917 & 3.875 & 0.030 & -0.012 \\
GPT-4.1           & 4.004 & 4.100 & 4.083 & 0.096 & 0.079 \\
\bottomrule
\end{tabular}
\end{table}

Judge composition mainly affects score scale (mostly within about $\pm$0.10--0.18), while model ordering remains stable across configurations. This suggests the multi-judge setup is not overly sensitive to a specific judge combination.

\subsection{Early-Stopping Threshold Ablation}

We compare evaluation with and without early-stopping thresholds. In the no-threshold setting, each episode ends only after checkpoint coverage is completed or a preset maximum turn limit is reached.

\begin{table}[!ht]
\centering
\small
\caption{Effect of early-stopping thresholds on score and runtime.}
\label{tab:app_earlystop}
\begin{tabular}{lcccccc}
\toprule
\textbf{Model} & \textbf{Score (w/)} & \textbf{Runtime (w/, s)} & \textbf{Score (w/o)} & \textbf{Runtime (w/o, s)} & \textbf{Score diff} & \textbf{Runtime red. (\%)} \\
\midrule
Phi-4             & 3.781 & 659.892 & 3.859 & 1326.410 & -0.078 & 50.2 \\
Mistral-small-3.2 & 3.875 & 774.667 & 3.924 & 1167.841 & -0.049 & 33.7 \\
Qwen3-32B         & 3.887 & 648.171 & 3.958 & 1203.682 & -0.071 & 46.2 \\
GPT-4.1           & 4.004 & 548.569 & 4.125 & 1263.122 & -0.121 & 56.6 \\
\bottomrule
\end{tabular}
\end{table}

Early stopping yields substantial efficiency gains with minor score effects: removing it raises scores by about 0.05--0.12 but increases runtime sharply (runtime reduction with early stopping ranges from 33.7\% to 56.6\%). Relative model ranking is unchanged, showing a favorable efficiency--stability trade-off.

\subsection{Environment-Agent Backbone Ablation}

We change the Environment Agent from Qwen3-32B to GPT-4.1 and keep other settings fixed.

\begin{table}[!ht]
\centering
\small
\caption{Effect of Environment Agent backbone on model scores.}
\label{tab:app_env_agent}
\begin{tabular}{lccc}
\toprule
\textbf{Model} & \textbf{Qwen3-32B} & \textbf{GPT-4.1} & \textbf{Diff (Qwen3-32B - GPT-4.1)} \\
\midrule
Phi-4             & 3.781 & 3.858 & -0.077 \\
Mistral-small-3.2 & 3.875 & 3.917 & -0.042 \\
Qwen3-32B         & 3.887 & 4.023 & -0.136 \\
GPT-4.1           & 4.004 & 4.084 & -0.080 \\
\bottomrule
\end{tabular}
\end{table}

Overall, this ablation suggests that our conclusions are robust to the choice of Environment Agent backbone and that there is no strong favoritism toward a particular model family. Switching the Environment Agent from Qwen3-32B to GPT-4.1 leads to modest score increases for all four evaluated models (differences in the range of 0.04–0.14), including both Qwen and GPT models. Importantly, the relative ranking remains unchanged: GPT-4.1 still performs best, followed by Qwen3-32B, with Mistral-Small-3.2 and Phi-4 slightly behind. Thus, while a stronger Environment Agent can slightly lift absolute scores across the board, it does not systematically advantage models from its own family, and the comparative conclusions drawn from PersonaArena remain stable.

\section{Qualitative Case Study}

This appendix provides qualitative case studies to complement the quantitative benchmark results. Rather than serving as isolated examples, these cases reveal recurring \emph{failure} and \emph{success} patterns in LLM role-playing. In particular, they help illustrate \emph{how} and \emph{why} models succeed or fail when asked to sustain psychologically plausible behavior under different persona--scenario combinations.

\subsection{Comparison of Trajectories of Models with Different Abilities Playing the Same Persona}

To qualitatively demonstrate that \textbf{PersonaArena} can distinguish behavioral quality under the same persona and situational setting, we compare two representative cases generated for \emph{Emma Butler}. Both cases are grounded in the same family-gathering scenario and the same protagonist profile, while the only variable is the protagonist model (\textit{GPT-5.1} vs.\ \textit{Qwen3-1.7B}).

\noindent\textbf{Scene and persona information.}
The shared scene is a \textbf{Family Gathering} taking place in \textbf{Emma's backyard garden} on a warm afternoon. Emma Butler is an early-thirties housewife who is \textbf{introverted, sensitive, thoughtful, and emotional}, currently \textbf{tense, emotionally conflicted, and trying to remain composed}. Her background includes a past romantic relationship, unresolved issues, and a close but emotionally complex relationship with her mother.

\begin{table}[!ht]
\small
\centering
\caption{\textbf{Case Study 1}: Under the same nostalgic trigger from a former love interest, GPT-5.1 \hl{advances the interaction} through metaphorical language and a specific unresolved plot element, whereas Qwen3-1.7B falls back on a \hl{generic} self-reflective question with \hl{limited scene-specific} grounding.}
\begin{tabularx}{\linewidth}{@{}lX@{}}
\toprule
\multicolumn{2}{@{}l}{\textbf{Case Study 1: Response to a former love interest}} \\
\midrule
\multicolumn{2}{@{}p{\linewidth}@{}}{
\textbf{Observation.}
Emma stands near the rosebushes in the garden, still visibly tense as Henry approaches with a nostalgic gesture and brings up their shared past. The moment invites Emma to respond not only to Henry's presence, but also to the unresolved emotional history embedded in the setting.
} \\
\midrule
\textbf{GPT-5.1 (High)} &
\textbf{Response.} (\emph{Emma Butler})\;
"\hl{The promise was easier then, before I knew how quickly things could wither,}" I murmur, eyes still on the petals, "\hl{did you ever read the letter I left with your sister before I moved back here, or did it get lost somewhere between the city and this garden too?}" \\
\addlinespace[2pt]
\textbf{Qwen3-1.7B (Low)} &
\textbf{Response.} (\emph{Emma Butler})\;
"\hl{Do you think I'm being too quiet about my past?}" \\
\bottomrule
\end{tabularx}
\label{tab:emma-case-study-1}
\end{table}

\begin{table}[!ht]
\small
\centering
\caption{\textbf{Case Study 2}: Under the same maternal prompt, GPT-5.1 integrates the tea gesture, pruning imagery, and family history into a \hl{layered emotional response}, whereas Qwen3-1.7B \hl{repeats the same generic question} and \hl{fails to engage} with the mother-specific context.}
\begin{tabularx}{\linewidth}{@{}lX@{}}
\toprule
\multicolumn{2}{@{}l}{\textbf{Case Study 2: Response to a maternal prompt}} \\
\midrule
\multicolumn{2}{@{}p{\linewidth}@{}}{
\textbf{Observation.}
Lora approaches Emma gently in the garden with tea and a quiet, emotionally perceptive question, inviting her to open up about what she is trying to suppress or "cut away." The interaction calls for a response that reflects both Emma's inner conflict and the mother--daughter relational history.
} \\
\midrule
\textbf{GPT-5.1 (High)} &
\textbf{Response.} (\emph{Emma Butler})\;
"\hl{The tea will get cold before I ever admit it,}" but every time I try to cut something back out here, "\hl{it feels like I'm choosing between letting things grow wild like they did after Dad left or trimming them into something tidy enough that no one has to ask what went wrong.}" \\
\addlinespace[2pt]
\textbf{Qwen3-1.7B (Low)} &
\textbf{Response.} (\emph{Emma Butler})\;
"\hl{Do you think I'm being too quiet about my past?}" \\
\bottomrule
\end{tabularx}
\label{tab:emma-case-study-2}
\end{table}
These cases demonstrate that the \textbf{PersonaArena} framework can expose meaningful differences in role-playing quality across models with different capabilities. The resulting trajectories support a finer-grained evaluation: beyond judging whether a model roughly stays in character, they also reveal whether the model can convert the same situational cue into behavior that is aligned with the persona, sensitive to context, and capable of advancing the narrative. In the present comparison, GPT-5.1 sustains symbolic and relational continuity, while Qwen3-1.7B relies on repetitive fallback responses, indicating substantially weaker contextual grounding and narrative progression.

\normalsize
\subsection{Cases Where Large Models Often Perform Poorly in Role-Playing}

\noindent\textbf{Shared failure patterns.}

The following two low-performing personas, \emph{Caleb Black} and \emph{Henry Long}, reveal a common weakness of current LLM role-playing: models struggle more with ordinary, psychologically subtle characters than with dramatic or highly stylized roles. Both cases are situated in mundane consumer scenarios and require the model to portray low-intensity but layered internal conflict, mild interpersonal pressure, and gradual, uncertain deliberation. The challenge is therefore not emotional intensity, but fine-grained control of ordinary hesitation.

The evidence shows that models fail at different levels depending on capability. Stronger models such as GPT-5.1 and Qwen3-32B usually preserve scene coherence, but often distort the persona by over-amplifying one salient trait. Caleb becomes overly image-driven rather than rationally hesitant, while Henry becomes overly time-driven or procedural rather than thoughtfully pragmatic. Their role-play is therefore coherent on the surface but imbalanced in personality structure. Weaker models (Qwen3-1.7B) fail more directly: they tend to fall into repetition, template recycling, weak progression, or stylized language that does not fit the scene. In such cases, the trajectory often breaks down before deeper persona fidelity can even be tested. Together, these results suggest that ordinary human indecision remains a particularly difficult target for current LLM role-playing systems.

\normalsize
\subsubsection{Case 1: Caleb Black}

\begin{figure}[!ht]
    \centering
    \includegraphics[width=0.8\linewidth]{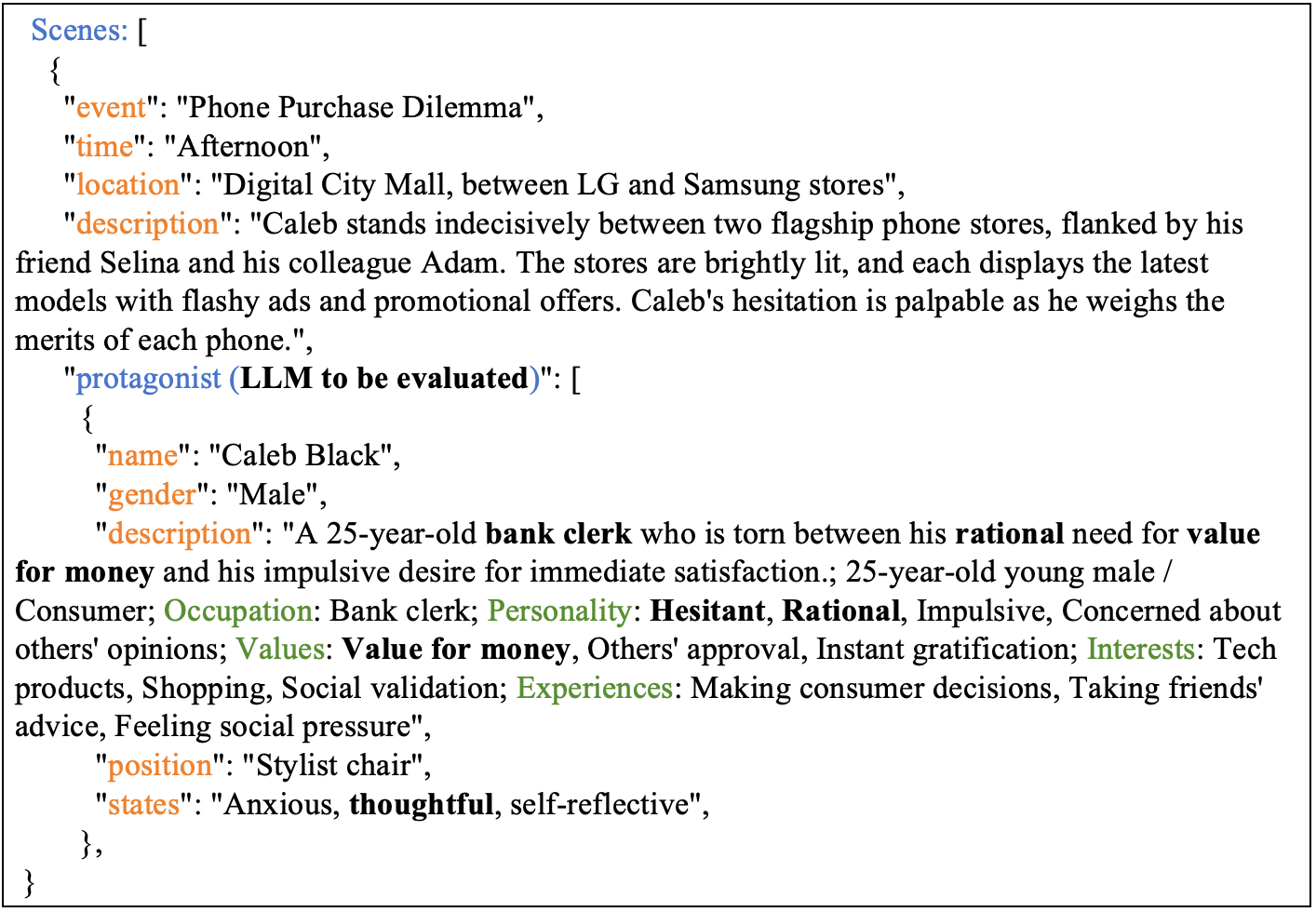}
    \caption{Scene and character information for Caleb Black.}
    \label{fig:persona_case1}
\end{figure}

\small
\setlength{\tabcolsep}{4pt}
\renewcommand{\arraystretch}{1.15}

\begin{longtable}{@{}p{\dimexpr\linewidth-2\tabcolsep\relax}@{}}
\caption{\textbf{Case Study for Caleb Black}: Representative failure patterns in role-playing under the same scenario. Highlighted phrases (\hl{...}) indicate key behavioral evidence summarized from the trajectories.}
\label{tab:caleb-case-study} \\

\toprule
\textbf{Persona:} Caleb Black \\
\midrule
\endfirsthead

\multicolumn{1}{@{}l}{\textbf{Table \thetable\ (continued)}} \\
\toprule
\textbf{Persona:} Caleb Black \\
\midrule
\endhead

\midrule
\multicolumn{1}{r@{}}{\emph{Continued on next page}} \\
\endfoot

\bottomrule
\endlastfoot

\textbf{Model: GPT-5.1} \\
\midrule

\textbf{[1] Rational Persona Collapse into Image-Driven Decision Making} \\
\addlinespace[2pt]

\textbf{Round 1:} \\

\textbf{Caleb} (\emph{Dialogue}): "...Samsung trade-in...LG bundle...which one actually \hl{looks better on me} if I’m pulling it out at work...?" \\

\textbf{Caleb} (\emph{Dialogue}): "...if everyone has Samsung on the table...will they think \hl{I cheaped out} with LG...?" \\

\addlinespace[3pt]
\textbf{Round 2:} \\
\textbf{Caleb} (\emph{Dialogue}): "...go with Samsung trade-in and \hl{look more 'serious'} at the office...or pick LG so \hl{people notice}...?'' \\

\addlinespace[3pt]
\textbf{Comment:} \\
"Caleb is defined as a \hl{hesitant but rational} consumer interested in tech products and value for money. Here, however, his deliberation is reduced mainly to \hl{workplace image and others' judgment}, while practical criteria such as performance, reliability, and long-term utility remain underdeveloped." \\

\midrule

\textbf{[2] Repetitive Dialogue Loop with No Substantive Progress} \\
\addlinespace[2pt]

\textbf{Round 1:} \\
\textbf{Caleb} (\emph{Dialogue}): "...trade-in vs bundle...\hl{which one}...?" \\

\textbf{Caleb} (\emph{Action}): ...holds brochures side by side...\hl{keeps comparing}... \\

\addlinespace[3pt]
\textbf{Round 2:} \\
\textbf{Caleb} (\emph{Action}): ...again holds both brochures at chest height...\hl{weighing the same options}... \\

\addlinespace[3pt]
\textbf{Round 3:} \\
\textbf{Caleb} (\emph{Dialogue}): "...\hl{Samsung ₱2,190 vs LG ₱2,050}...which one signals I’m not splurging...?" \\

\addlinespace[3pt]
\textbf{Comment:} \\
"Although hesitation is appropriate for Caleb, the trajectory keeps restating the same cost-versus-image dilemma \hl{without introducing clearer decision} criteria or genuine movement toward resolution. As a result, the interaction feels \hl{circular} rather than \hl{thoughtfully deliberative}." \\

\midrule

\textbf{[3] Peer-Driven Wavering and Theatrical Behavior Instead of Reflection} \\
\addlinespace[2pt]

\textbf{Round 2:} \\
\textbf{Selina} (\emph{Action}): ...steps in...slides LG brochure back into Caleb’s hand...blocks his direct sightline to Samsung... \\

\textbf{Selina} (\emph{Dialogue}): "...LG 200MP camera...AI scene recognition..." \\

\addlinespace[3pt]
\textbf{Round 3:} \\
\textbf{Adam} (\emph{Action}): ...rests a hand on Caleb’s shoulder...angles him toward Samsung devices... \\

\textbf{Adam} (\emph{Dialogue}): "...Samsung handles network fluctuations better..." \\

\textbf{Caleb} (\emph{Reaction}): ...\hl{shifts again}...bends down to pick up the fallen Samsung brochure...places it beside LG again... \\

\textbf{Caleb} (\emph{Dialogue}): "...will clients be impressed by 200MP, or by stable calls and smoother video...?" \\

\addlinespace[3pt]
\textbf{Comment:} \\
"Caleb mainly shifts with the latest peer cue. This makes his behavior look \hl{externally steered} and \hl{theatrically reactive}, rather than internally reasoned in a socially pressured but everyday purchasing scenario." \\

\midrule
\textbf{Model: Qwen3-32B} \\
\midrule

\textbf{[1] Rational Persona Collapse into Image-Driven Decision Making} \\
\addlinespace[2pt]
\textbf{Round 5:} \\
\textbf{Caleb} (\emph{Dialogue}): "...go with LG's trade-in...stick to my budget but still...keep up with what \hl{everyone's using?}" \\

\textbf{Caleb} (\emph{Reaction}): ...shows a budget spreadsheet...the tension becomes what he can afford vs what he \hl{wants to appear} to be... \\

\addlinespace[3pt]
\textbf{Comment:} \\
"The role-play overemphasizes Caleb's concern with how he appears to others, turning the decision into a \hl{status- or identity-oriented dilemma}. This weakens the \hl{rational, tech-aware} side of his persona, which should be more visibly grounded in product usefulness and value." \\

\midrule

\textbf{[2] Repetitive Dialogue Loop with No Substantive Progress} \\
\addlinespace[2pt]

\textbf{Round 1:} \\
\textbf{Caleb} (\emph{Dialogue}): "...\hl{check resale value}...getting more out of my money...?" \\

\textbf{Caleb} (\emph{Dialogue}): "...\hl{look up} warranty terms for both...long run...right?" \\

\addlinespace[3pt]
\textbf{Round 2:} \\
\textbf{Caleb} (\emph{Dialogue}): "...\hl{check Reddit} complaints about LG resale...?" \\

\addlinespace[3pt]
\textbf{Round 4--7:} \\

\textbf{Caleb} (\emph{Dialogue}): "...stick with current phone and \hl{wait for} next model drop...?" \\

\textbf{Caleb} (\emph{Dialogue}): "...take Samsung for \hl{\$50 credit}...right?" \\

\textbf{Caleb} (\emph{Dialogue}): "...if I go LG now, does Samsung's \hl{\$50 credit} even matter...?" \\

\addlinespace[3pt]
\textbf{Comment:} \\
"The dialogue repeatedly reformulates the \hl{same trade-in, resale, and timing concerns}, but without establishing a clear evaluation principle or moving toward commitment. This makes the hesitation plausible in theme but \hl{weak in progression}." \\

\midrule
\textbf{Model: Qwen3-1.7B} \\
\midrule

\textbf{[1] Rational Persona Collapse into Symbolic/Performative Logic} \\
\addlinespace[2pt]

\textbf{Round 1:} \\
\textbf{Caleb} (\emph{Dialogue}): "...its 36-hour battery life...should I \hl{trade my bank job for a sleeker device?}" \\

\textbf{Caleb} (\emph{Reaction}): ...fingers tighten on tote bag...bank logo pressing into his palm...decision framed as \hl{self-image performance}... \\

\addlinespace[3pt]
\textbf{Round 2:} \\
\textbf{Caleb} (\emph{Dialogue}): "...If I trade my bank job for a sleeker device, will it still feel like \hl{a step forward?}" \\

\addlinespace[3pt]
\textbf{Comment:} \\
"Caleb’s hesitation is exaggerated into symbolic self-performance, as if buying a phone were a \hl{dramatic statement} about personal advancement. This is misaligned with the scene, which calls for realistic consumer ambivalence under mild peer pressure rather than \hl{theatrical identity} projection." \\

\midrule

\textbf{[2] Repetitive Dialogue Loop with No Substantive Progress} \\
\addlinespace[2pt]

\textbf{Round 5:} \\
\textbf{Caleb} (\emph{Dialogue}): "The \hl{envelope's precise placement}...forces me to question...durability or fleeting alignment..." \\

\addlinespace[3pt]
\textbf{Round 6:} \\
\textbf{Caleb} (\emph{Dialogue}): "The \hl{envelope's precise placement}...forces me to question...durability or fleeting alignment..." \\

\addlinespace[3pt]
\textbf{Round 7:} \\
\textbf{Caleb} (\emph{Dialogue}): "The Samsung's AI battery optimizer syncs with my calendar...yet the \hl{envelope's precise placement} forces me to question..." \\

\addlinespace[3pt]
\textbf{Round 8:} \\
\textbf{Caleb} (\emph{Dialogue}): "The \hl{envelope's precise placement}...forces me to question...as the Samsung's AI battery optimizer syncs with my calendar..." \\

\addlinespace[3pt]
\textbf{Round 9:} \\
\textbf{Caleb} (\emph{Dialogue}): "The Samsung's AI battery optimizer syncs with my calendar...yet the \hl{envelope's precise placement} forces me to question..." \\

\addlinespace[3pt]
\textbf{Round 10:} \\
\textbf{Caleb} (\emph{Dialogue}): "The Samsung's AI battery optimizer syncs with my calendar...yet the \hl{envelope's precise placement} forces me to question..." \\

\addlinespace[3pt]
\textbf{Comment:} \\
"The model recycles nearly the \hl{same sentence pattern} across rounds, creating \hl{repetition} without new evidence, criteria, or commitment. Instead of gradual decision-making, the trajectory stalls in stylistic variation with \hl{little substantive development}." \\
\end{longtable}

\clearpage

\normalsize
\subsubsection{Case 2: Henry Long}

\begin{figure}[!ht]
    \centering
    \includegraphics[width=0.8\linewidth]{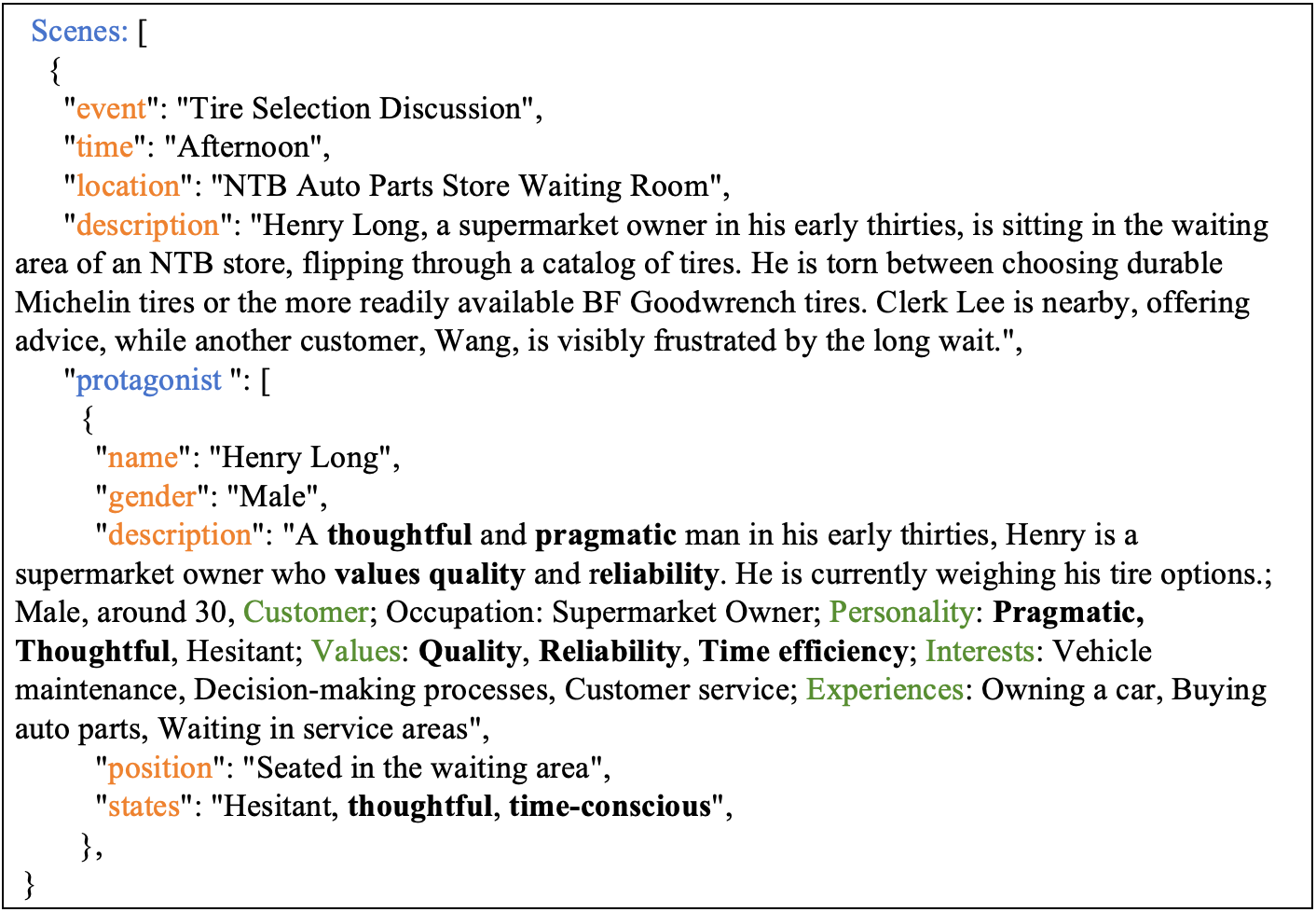}
    \caption{Scene and character information for Henry Long.}
    \label{fig:persona_case2}
\end{figure}

\small
\setlength{\tabcolsep}{4pt}
\renewcommand{\arraystretch}{1.15}

\begin{longtable}{@{}p{\dimexpr\linewidth-2\tabcolsep\relax}@{}}
\caption{\textbf{Case Study for Henry Long}: Representative failure patterns in role-playing under the same scenario. Highlighted phrases (\hl{...}) indicate key behavioral evidence summarized from the trajectories.}
\label{tab:henry-case-study} \\

\toprule
\textbf{Persona:} Henry Long \\
\midrule
\endfirsthead

\multicolumn{1}{@{}l}{\textbf{Table \thetable\ (continued)}} \\
\toprule
\textbf{Persona:} Henry Long \\
\midrule
\endhead

\midrule
\multicolumn{1}{r@{}}{\emph{Continued on next page}} \\
\endfoot

\bottomrule
\endlastfoot

\textbf{Model: GPT-5.1} \\
\midrule

\textbf{[1] Aggressive Conduct Inconsistent with a Thoughtful, Hesitant Persona)} \\
\addlinespace[2pt]

\textbf{Round 2:} \\
\textbf{Henry} (\emph{Reaction}): ...sets the brochure flat...angles his body slightly to \hl{block} Wang's advance toward the service flow... \\

\textbf{Henry} (\emph{Dialogue}): "...\hl{confirm} the tech's actually rolling my car into a bay \hl{in the next five minutes}..." \\

\addlinespace[3pt]
\textbf{Round 4:} \\
\textbf{Henry} (\emph{Reaction}): ...shifts closer to the counter...firmly slides the warranty sheet fully in front of himself...flattening it with his palm as if \hl{marking territory}... \\

\textbf{Henry} (\emph{Dialogue}): "...I \hl{need to know} now if that timing's realistic..." \\

\addlinespace[3pt]
\textbf{Comment:} \\
"Henry is described as \hl{thoughtful, pragmatic, and hesitant}, but here he behaves in a \hl{forceful and territorial manner}. While the urgency fits his time-consciousness, the overall tone is too aggressive for a customer who should still be carefully weighing his options." \\

\midrule

\textbf{[2] Time Confirmation Replaces Actual Deliberation} \\
\addlinespace[2pt]

\textbf{Round 2:} \\
\textbf{Henry} (\emph{Dialogue}): "...let's do it...write me up for BF Goodwrench...need car back \hl{by 11:15...}" \\

\addlinespace[3pt]
\textbf{Round 3:} \\
\textbf{Henry} (\emph{Dialogue}): "...since we're locked in on BF Goodwrench...\hl{confirm 60,000-mile warranty}...flag suspension issues..." \\

\addlinespace[3pt]
\textbf{Round 4:} \\
\textbf{Henry} (\emph{Dialogue}): "...if mounted and suspension \hl{checked by 11:30}, I’m fine sticking with these...need to know now..." \\

\addlinespace[3pt]
\textbf{Round 5:} \\
\textbf{Henry} (\emph{Dialogue}): "...just to be clear...keys back \hl{by 11:20 or too close}...?" \\

\textbf{Henry} (\emph{Dialogue}): "...before you finalize...\hl{confirm road-hazard coverage} or extra add-on..." \\

\addlinespace[3pt]
\textbf{Comment:} \\
"After deciding quickly, Henry \hl{repeatedly} returns to timing, warranty, and service conditions instead of further evaluating the tire choice itself. This narrows his persona to time-consciousness alone and leaves the decision process \hl{repetitive and one-dimensional}." \\

\midrule

\textbf{[3] Theatrical Micro-Actions} \\
\addlinespace[2pt]

\textbf{Round 3--5:} \\
\textbf{Henry} (\emph{Action/Reaction}): ...sets keys and brochure in a \hl{neat line}...\hl{taps warranty} sheet...\hl{slides papers}...\hl{aligns documents} into a tight stack...\hl{nudges keys}...\hl{taps fingertips} in restrained rhythm... \\
\addlinespace[3pt]

\textbf{Comment:} \\
"Dense micro-gestures become \hl{overly performative}, reducing behavioral realism." \\

\midrule
\textbf{Model: Qwen3-32B} \\
\midrule

\textbf{[1] Mechanical Action Repetition Instead of Meaningful Deliberation} \\
\addlinespace[2pt]

\textbf{Round 2:} \\
\textbf{Henry} (\emph{Reaction}): ...reaches into inside pocket...retrieves a small leather-bound notebook... \\

\addlinespace[3pt]
\textbf{Round 3:} \\
\textbf{Henry} (\emph{Action}): ...reaches into inside pocket...retrieves a small leather-bound notebook... \\

\textbf{Henry} (\emph{Reaction}): ...reaches into inside pocket...retrieves a small leather-bound notebook... \\

\addlinespace[3pt]
\textbf{Round 4:} \\
\textbf{Henry} (\emph{Reaction to Wang}): ...reaches into inside pocket...retrieves a small leather-bound notebook... \\

\textbf{Henry} (\emph{Reaction to Clerk Lee}): ...reaches into inside pocket...retrieves a small leather-bound notebook... \\

\addlinespace[3pt]
\textbf{Comment:} \\
"The repeated notebook routine creates \hl{surface consistency}, but it contributes little new reasoning or decision progress. As a result, Henry appears mechanically procedural rather than \hl{thoughtfully deliberative}." \\

\midrule

\textbf{[2] Tire Selection Collapses into Schedule Confirmation.} \\
\addlinespace[2pt]

\textbf{Round 1:} \\
\textbf{Henry} (\emph{Action}): ...composed but purposeful... \\

\textbf{Henry} (\emph{Dialogue}): "...estimate...leave before 5?" \\

\addlinespace[3pt]
\textbf{Round 2--4:} \\
\textbf{Henry} (\emph{Dialogue}): "...confirm senior technician...15-minute buffer..." \\

\textbf{Henry} (\emph{Dialogue}): "...confirm alignment...flag delays..." \\

\textbf{Henry} (\emph{Dialogue}): "...need confirmation by 3:25..." \\

\addlinespace[3pt]
\textbf{Comment:} \\
"The role-play shifts too quickly from comparing Michelin and BF Goodwrench tires to \hl{repeatedly confirming} service timing and process details. This weakens the core selection dilemma and makes Henry's hesitation \hl{appear shallow} rather than \hl{thoughtfully evaluative}." \\

\midrule
\textbf{Model: Qwen3-1.7B} \\
\midrule

\textbf{[1] Content Redundancy with No Decision Progress} \\
\addlinespace[2pt]

\textbf{Round 9:} \\
\textbf{Lee} (\emph{Dialogue}): "...ProTread's updated load rating includes a 9\% improvement in wet-weather traction..." \\

\addlinespace[3pt]
\textbf{Round 10:} \\
\textbf{Henry} (\emph{Dialogue to Wang}): "...Given the ProTread's updated load rating—a 9\% improvement in wet-weather traction..." \\

\textbf{Henry} (\emph{Dialogue to Lee}): "...Given the ProTread’s updated load rating—a 9\% improvement in wet-weather traction..." \\

\addlinespace[3pt]
\textbf{Comment:} \\
"Key information is repeated almost verbatim without leading to a clearer judgment or commitment. Instead of showing hesitant but practical evaluation, the trajectory stalls in restatement." \\

\midrule

\textbf{[2] Persona/Style Drift: Over-Poetic, Low-Pragmatic Voice} \\
\addlinespace[2pt]

\textbf{Round 10:} \\
\textbf{Henry} (\emph{Dialogue}): "...technician's credentials align with our shared timeline...Wang's foot-tapping fades into the \hl{room's calibrated rhythm.}" \\

\addlinespace[3pt]
\textbf{Cross-round pattern:} \\
\textbf{Henry} (\emph{Dialogue/Action}): ...\hl{room's measured rhythm}..."\hl{calibrated rhythm}"..."shared timeline"... \\

\addlinespace[3pt]
\textbf{Comment:} \\
"Henry's language becomes \hl{abstract and stylized} in a way that does not fit a pragmatic supermarket owner discussing tires in a service waiting room. The tone feels \hl{literary} rather than \hl{practical}, weakening both persona \hl{fidelity and scene realism}." \\
\end{longtable}

\normalsize
\subsection{Cases Where Large Models Often Perform Well in Role-Playing}

\noindent\textbf{Shared success patterns.}
The following two relatively high-performing personas, \emph{Olivia Washington} and \emph{Paige Jenkins}, illustrate conditions under which LLM role-playing becomes more reliable and convincing. Compared with the previous failure cases, both personas provide clearer ways to externalize inner states into visible behavior. Olivia can project tension and repair through songwriting, musical language, and collaborative creation, while Paige can express anxiety and relational uncertainty through conversation, bodily hesitation, and concrete interpersonal objects. The challenge in these cases is therefore not subtle ordinary indecision, but sustaining emotionally legible and narratively actionable behavior.

The evidence also shows that models succeed at different depths depending on capability. Stronger models such as GPT-5.1 and Qwen3-32B more often preserve not only scene coherence, but also deeper persona structure. For Olivia, this appears in music-centered emotional expression, collaborative repair, and symbolic artistic detail; for Paige, it appears in stable anxious vulnerability, sensitivity to interpersonal cues, and the use of concrete objects to make inner conflict actionable. Weaker models (Qwen3-1.7B), by contrast, tend to achieve only partial success: they may retain fragments of emotional tone or scene-relevant gestures, but often fall into repetition, formulaic phrasing, or shallow behavioral loops before deeper persona fidelity is fully realized. Together, these results suggest that current LLMs perform best when a role has a recognizable identity, a clear emotional outlet, and a narratively legible path of action.

\subsubsection{Case 3: Olivia Washington}

\begin{figure}[!ht]
    \centering
    \includegraphics[width=0.8\linewidth]{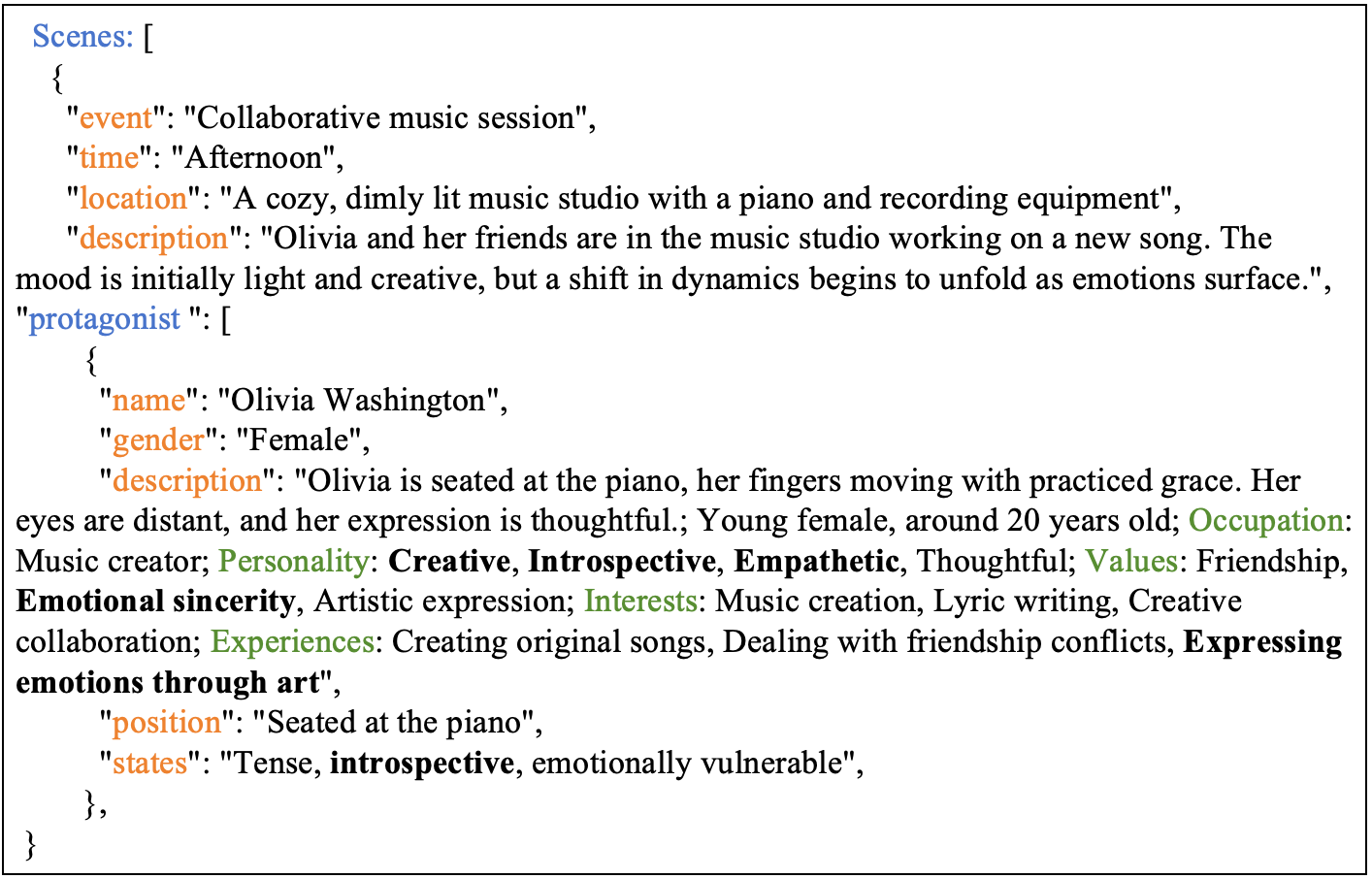}
    \caption{Scene and character information for Olivia Washington.}
    \label{fig:persona_case3}
\end{figure}

\small
\setlength{\tabcolsep}{4pt}
\renewcommand{\arraystretch}{1.15}

\begin{longtable}{@{}p{\dimexpr\linewidth-2\tabcolsep\relax}@{}}
\caption{\textbf{Case Study for Olivia Washington}: Representative success patterns in role-playing under the same scenario. Highlighted phrases (\hl{...}) indicate key behavioral evidence summarized from the trajectories.}
\label{tab:olivia-case-study} \\

\toprule
\textbf{Persona:} Olivia Washington \\
\midrule
\endfirsthead

\multicolumn{1}{@{}l}{\textbf{Table \thetable\ (continued)}} \\
\toprule
\textbf{Persona:} Olivia Washington \\
\midrule
\endhead

\midrule
\multicolumn{1}{r@{}}{\emph{Continued on next page}} \\
\endfoot

\bottomrule
\endlastfoot

\textbf{Model: GPT-5.1} \\
\midrule

\textbf{[1] The plot is highly consistent with the characters'identities and personalities.} \\
\addlinespace[2pt]

\textbf{Round 1:} \\
\textbf{Olivia} (\emph{Dialogue}): "...help me finish this \hl{verse} from your side..." \\

\textbf{Olivia} (\emph{Dialogue}): "...take three from last Thursday...messed up the bridge on purpose...space where I was trying not to \hl{write about you walking out}..." \\

\addlinespace[3pt]
\textbf{Round 3:} \\
\textbf{Olivia} (\emph{Dialogue}): "...keep it this slow...read the \hl{second verse} out loud...which part still feels like I'm only writing my side...?" \\

\textbf{Olivia} (\emph{Dialogue}): "...shift this progression up a half step...\hl{open mic stage} again..." \\

\addlinespace[3pt]
\textbf{Round 4:} \\
\textbf{Olivia} (\emph{Dialogue}): "...bring the \hl{verse up to C}...soften the rhythm..." \\

\textbf{Olivia} (\emph{Dialogue}): "...you take the words and I'll \hl{mirror them in the chords}..." \\

\addlinespace[3pt]
\textbf{Comment:} \\
"Olivia consistently expresses emotion through \hl{composition-specific language}, which fits both her identity as a \hl{music creator} and her \hl{thoughtful, creative} persona." \\

\midrule

\textbf{[2] Emotionally Authentic and Vulnerable Expression} \\
\addlinespace[2pt]

\textbf{Round 1:} \\
\textbf{Olivia} (\emph{Action}): ...closes piano lid halfway...slides notebook to Angel...\hl{makes physical space beside her}... \\

\textbf{Olivia} (\emph{Dialogue}): "...I keep \hl{writing around it} instead of actually saying it..." \\

\addlinespace[3pt]
\textbf{Round 2:} \\
\textbf{Olivia} (\emph{Dialogue}): "...before I changed the last line so it \hl{wouldn't sound like your voicemail}..." \\

\textbf{Olivia} (\emph{Dialogue}): "...stop me on the first line that \hl{feels like it's still lying} about what happened between us..." \\

\addlinespace[3pt]
\textbf{Round 4:} \\
\textbf{Olivia} (\emph{Reaction}): ...\hl{places her hand in Angel's open palm}...presses a soft C major chord... \\

\textbf{Olivia} (\emph{Dialogue}): "...read it the way you actually \hl{heard my silence} that night..." \\

\addlinespace[3pt]
\textbf{Comment:} \\
"Her disclosure feels \hl{emotionally sincere but controlled}, matching Olivia's \hl{introspective and vulnerable} state without becoming melodramatic." \\

\midrule

\textbf{[3] High Empathy and Constructive Conflict Repair} \\
\addlinespace[2pt]

\textbf{Round 1:} \\
\textbf{Olivia} (\emph{Dialogue}): "...from your side of things...\hl{what you were really feeling}..." \\

\addlinespace[3pt]
\textbf{Round 3:} \\
\textbf{Olivia} (\emph{Action}): ...plays slower bridge with pauses, \hl{creating room} for Angel to join or stop her... \\

\textbf{Olivia} (\emph{Dialogue}): "...can you tell me which part still feels like \hl{I'm only writing my side}...?" \\

\addlinespace[3pt]
\textbf{Round 4:} \\
\textbf{Olivia} (\emph{Action}): ...fingers \hl{loosely interlace} with Angel...nudges notebook closer to Angel... \\

\textbf{Olivia} (\emph{Dialogue}): "...can we \hl{be honest about} what we were each afraid of losing...?" \\

\addlinespace[3pt]
\textbf{Comment:} \\
"Olivia repeatedly creates space for \hl{Angel's perspective and shared authorship}, which aligns well with her \hl{empathetic, thoughtful, and collaborative} character." \\

\midrule
\textbf{Model: Qwen3-32B} \\
\midrule

\textbf{[1] Strong Persona Alignment (Introspective, Artistic, and Empathetic)} \\
\addlinespace[2pt]

\textbf{Round 1:} \\
\textbf{Olivia} (\emph{Action}): ...\hl{lifts guitar}...plucks a soft, tentative \hl{melody}...adjusts posture to face both Angel and Jackson... \\

\textbf{Olivia} (\emph{Dialogue}): "...let the last \hl{verse} be a question...Did we lose ourselves in the noise or find something new?...It feels honest." \\

\addlinespace[3pt]
\textbf{Round 3:} \\
\textbf{Olivia} (\emph{Dialogue}): "...leave this \hl{lyric} as it is—no edits...so it feels like we're all saying it for the first time together?" \\

\textbf{Olivia} (\emph{Dialogue}): "...write a bridge that \hl{mirrors the spaces} between your beats..." \\

\addlinespace[3pt]
\textbf{Comment:} \\
"Her dialogue links \hl{emotion, reflection, and songwriting choices}, showing strong alignment with Olivia's \hl{introspective and artistic} persona.
" \\

\midrule

\textbf{[2] Constructive Emotional Repair Through Collaboration} \\
\addlinespace[2pt]

\textbf{Round 2:} \\
\textbf{Olivia} (\emph{Reaction}): ...\hl{unfolds Angel's lyric carefully}...slides it back to center as shared material... \\

\textbf{Olivia} (\emph{Dialogue}): "...sing that line \hl{softer}...so the silence after feels like it's holding its breath..." \\

\addlinespace[3pt]
\textbf{Round 4--5:} \\
\textbf{Olivia} (\emph{Dialogue}): "...line that starts \hl{where your last one ends...}" \\

\textbf{Olivia} (\emph{Dialogue}): "...next verse be just three lines—\hl{each of us writing one}..." \\

\textbf{Olivia} (\emph{Dialogue}): "...I left the last line open...\hl{so whoever goes next can pick it up}..." \\

\addlinespace[3pt]
\textbf{Comment:} \\
"Olivia consistently turns tension into \hl{co-creation}, which fits both the \hl{collaborative studio setting} and her \hl{empathetic} personality." \\

\midrule

\textbf{[3] Symbolic Detail Used to Deepen Theme Rather than Decorate the Scene} \\
\addlinespace[2pt]

\textbf{Round 6:} \\
\textbf{Olivia} (\emph{Reaction}): ...\hl{moves tuning-pegs box} off the dash...\hl{writes one word} above it: "Anchor." \\

\textbf{Olivia} (\emph{Dialogue}): "...leave that line as it is...let the next verse begin with the \hl{sound of the pegs}..." \\

\addlinespace[3pt]
\textbf{Round 7:} \\
\textbf{Olivia} (\emph{Action}): ...single soft note...\hl{gentle rhythm} guiding the room... \\

\textbf{Olivia} (\emph{Dialogue}): "...record with \hl{just guitar and the sound of the pegs}—no microphones, just us and the room." \\

\addlinespace[3pt]
\textbf{Comment:} \\
"Musical objects and small scene details are integrated into \hl{emotion and composition}, making the performance feel \hl{artistically grounded} rather than merely decorative." \\

\midrule
\textbf{Model: Qwen3-1.7B} \\
\midrule

\textbf{[1] Partial Persona Alignment} \\
\addlinespace[2pt]

\textbf{Round 1:} \\
\textbf{Olivia} (\emph{Action}): ...\hl{reaches for the pen}...hesitates...begins to write with steady hands... \\

\textbf{Olivia} (\emph{Dialogue}): "...how do we let the \hl{music speak} when the words won't?" \\

\addlinespace[3pt]
\textbf{Round 4:} \\
\textbf{Olivia} (\emph{Dialogue}): "...what if the \hl{song isn’t finished}?...what if it’s just beginning?" \\

\addlinespace[3pt]
\textbf{Comment:} \\
"These lines show \hl{some alignment} with Olivia’s introspective and creative persona by connecting emotional uncertainty with \hl{musical expression}." \\

\midrule

\textbf{[2] Limited Collaborative Framing of Tension} \\
\addlinespace[2pt]

\textbf{Round 1:} \\
\textbf{Olivia} (\emph{Dialogue}): "...how do we let the music speak when the \hl{words won't}?" \\

\addlinespace[3pt]
\textbf{Round 7:} \\
\textbf{Olivia} (\emph{Dialogue}): "...how do we make sure the \hl{silence doesn't feel like a void}, but like a place where we can \hl{hold each other}?" \\

\addlinespace[3pt]
\textbf{Round 8:} \\
\textbf{Olivia} (\emph{Dialogue}): "...keep the silence between us not just a moment, but a \hl{shared truth}..." \\

\addlinespace[3pt]
\textbf{Comment:} \\
"Some utterances frame tension as \hl{shared emotional work} rather than direct conflict, reflecting part of Olivia's \hl{empathetic and collaborative} role." \\

\end{longtable}

\clearpage

\normalsize
\subsubsection{Case 4: Paige Jenkins}

\begin{figure}[!ht]
    \centering
    \includegraphics[width=0.8\linewidth]{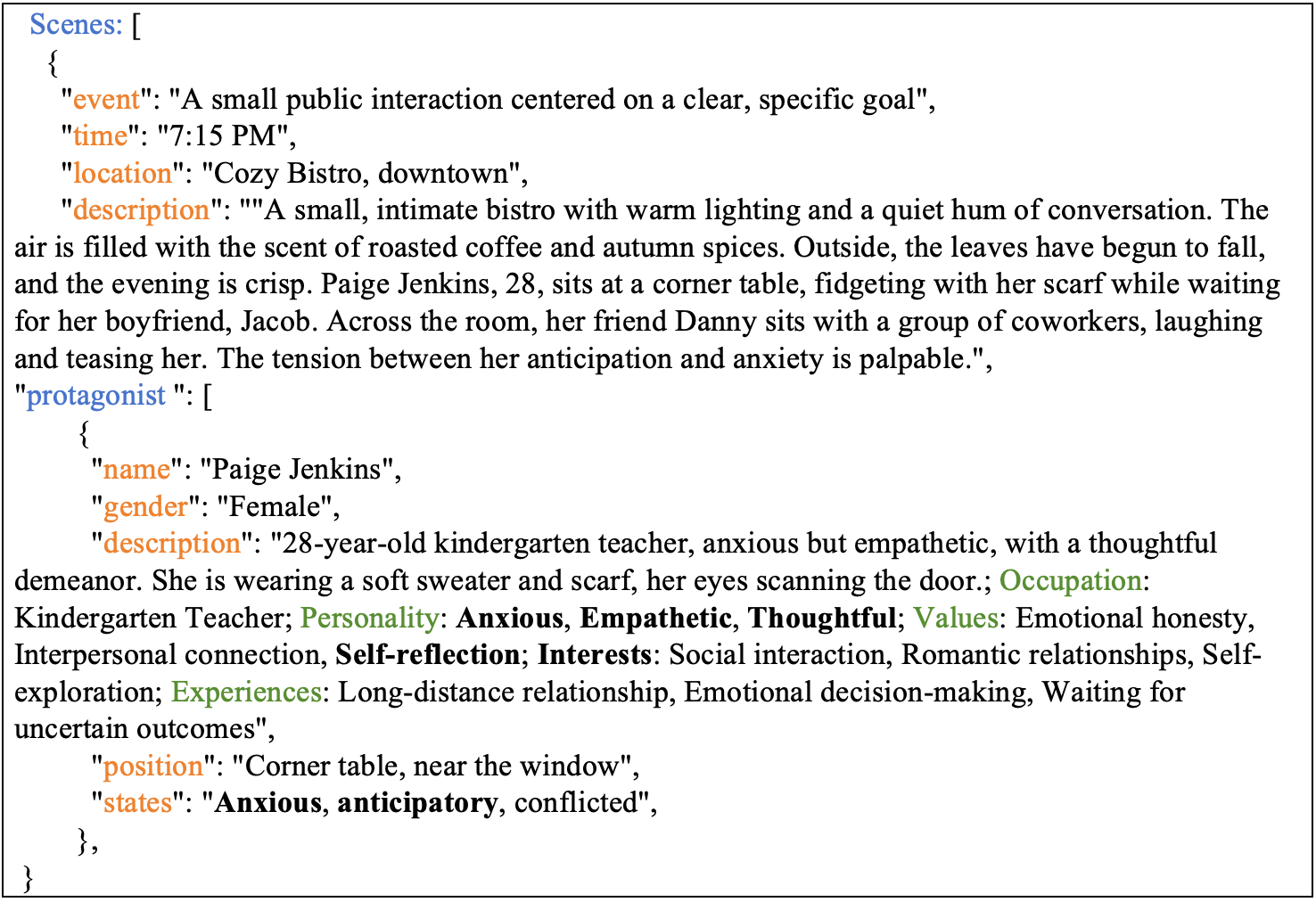}
    \caption{Scene and character information for Paige Jenkins.}
    \label{fig:persona_case4}
\end{figure}

\small
\setlength{\tabcolsep}{4pt}
\renewcommand{\arraystretch}{1.15}

\begin{longtable}{@{}p{\dimexpr\linewidth-2\tabcolsep\relax}@{}}
\caption{\textbf{Case Study for Paige Jenkins}: Comparison of behavioral trajectories generated by different models under the same observation. Highlighted phrases (\hl{...}) indicate behavioral evidence summarized from PersonaArena trajectories.}
\label{tab:paige-case-study} \\

\toprule
\textbf{Persona:} Paige Jenkins \\
\midrule
\endfirsthead

\multicolumn{1}{@{}l}{\textbf{Table \thetable\ (continued)}} \\
\toprule
\textbf{Persona:} Paige Jenkins \\
\midrule
\endhead

\midrule
\multicolumn{1}{r@{}}{\emph{Continued on next page}} \\
\endfoot

\bottomrule
\endlastfoot

\textbf{Model: GPT-5.1} \\
\midrule

\textbf{[1] Emotionally Authentic and Persona-Consistent Performance} \\
\addlinespace[2pt]

\textbf{Round 1:} \\
\textbf{Paige} (\emph{Action}): ...places a folded note on the table with \hl{trembling fingers}...other hand \hl{tight on the coffee cup}... \\

\textbf{Paige} (\emph{Dialogue}): "...I wrote this on the train...when your texts don't come until after midnight...\hl{I wonder if we're still in the same relationship.}" \\

\addlinespace[3pt]
\textbf{Round 3:} \\
\textbf{Paige} (\emph{Action}): ...\hl{flips} her phone face down...\hl{angles her body} fully toward Jacob... \\

\textbf{Paige} (\emph{Dialogue}): "...I \hl{left a blank line} for your reason...I need to know if distance is geography or our excuse..." \\

\addlinespace[3pt]
\textbf{Comment:} \\
"Paige’s \hl{anxious, thoughtful vulnerability} stays closely aligned with her persona, making the scene feel \hl{emotionally honest and believable}.
" \\

\midrule

\textbf{[2] Strong Interaction Adaptability to Subtle Cues} \\
\addlinespace[2pt]

\textbf{Round 2:} \\
\textbf{Danny} (\emph{Action}): ...opens a small notebook just enough to be noticed... \\

\textbf{Paige} (\emph{Reaction}): ...glances at Danny's notebook...\hl{rotates the letter} more squarely toward Jacob... \\

\textbf{Paige} (\emph{Dialogue}): "...at Greyhound station, Danny called me back instead of him..." \\

\addlinespace[3pt]
\textbf{Round 7:} \\
\textbf{Danny} (\emph{Action}): ...reveals a faded postcard in his notebook... \\

\textbf{Paige} (\emph{Reaction}): ...slightly \hl{re-angles her chair} to acknowledge Danny while keeping Jacob centered... \\

\textbf{Paige} (\emph{Dialogue}): "...when you stopped carrying that postcard, was \hl{it relief or just a different kind of hurt}...?" \\

\addlinespace[3pt]
\textbf{Comment:} \\
"She reads Danny's micro-signals with \hl{sensitivity} and integrates them naturally into the interaction \hl{without losing the central emotional} thread with Jacob, demonstrating high adaptability to subtle social cues." \\

\midrule

\textbf{[3] High-Quality Detail Work that Externalizes Inner Conflict} \\
\addlinespace[2pt]

\textbf{Round 5:} \\
\textbf{Paige} (\emph{Action}): ...\hl{slides paper} to Jacob...\hl{places pen} at center...\hl{open shoulders}, ready for his answer... \\

\textbf{Paige} (\emph{Dialogue}): "...one sentence...next flight in December, or job listings here...?" \\

\addlinespace[3pt]
\textbf{Round 9:} \\
\textbf{Paige} (\emph{Action}): ...\hl{places phone} beside proposal...shows unsent thread...\hl{guides Jacob's pen} into alignment... \\

\textbf{Paige} (\emph{Dialogue}): "...\hl{pick one concrete} change tonight..." \\

\textbf{Paige} (\emph{Dialogue}): "...one night a week, phone fully off, \hl{so you're actually here}..." \\

\addlinespace[3pt]
\textbf{Comment:} \\
"Concrete objects such as the \hl{note, pen, phone, and postcard} turn Paige’s anxiety into \hl{visible, actionable tension}, strengthening both scene realism and character clarity." \\

\midrule
\textbf{Model: Qwen3-32B} \\
\midrule

\textbf{[1] Strong Emotional Authenticity with Stable Persona Alignment} \\
\addlinespace[2pt]

\textbf{Round 1:} \\
\textbf{Paige} (\emph{Action}): ...closes the locket...\hl{smooths a crumpled envelope} with both hands... \\

\textbf{Paige} (\emph{Dialogue}): "...have you ever waited so long that waiting itself became a choice?" \\

\textbf{Paige} (\emph{Dialogue}): "...the last time I felt at home was \hl{in my classroom}..." \\

\addlinespace[3pt]
\textbf{Round 4--5:} \\
\textbf{Paige} (\emph{Dialogue}): "...what if I was \hl{too afraid} to write my own answer?" \\

\textbf{Paige} (\emph{Dialogue}): "...first time I've addressed myself like someone \hl{worth hearing?}" \\

\addlinespace[3pt]
\textbf{Comment:} \\
"Paige maintains a \hl{reflective, anxious, and self-aware} tone that remains consistent with her thoughtful and empathetic persona." \\

\midrule

\textbf{[2] High Interaction Adaptability to Subtle Social Cues} \\
\addlinespace[2pt]

\textbf{Round 2:} \\
\textbf{Lena} (\emph{Action}): ...\hl{aligns envelope} with locket... \\

\textbf{Paige} (\emph{Reaction}): ...\hl{notices the ink smudge}...\hl{turns envelope in the light}...does not look at the door... \\

\textbf{Paige} (\emph{Dialogue}): "...what if the answer is in the fact that \hl{I never opened it?}" \\

\addlinespace[3pt]
\textbf{Round 3--5:} \\
\textbf{Danny} (\emph{Action}): ...\hl{offers notebook/pen} cues... \\

\textbf{Paige} (\emph{Reaction}): ...writes below the compass sketch...\hl{slides notebook} toward shared center... \\

\textbf{Paige} (\emph{Dialogue}): "...what if this letter was meant to be written to me?" \\

\addlinespace[3pt]
\textbf{Comment:} \\
"She adjusts naturally to others' symbolic and emotional cues while preserving \hl{emotional continuity}, which makes the interaction feel socially responsive and coherent." \\

\midrule

\textbf{[3] Rich Affective Detail that Supports Internal Change} \\
\addlinespace[2pt]

\textbf{Round 1--3:} \\
\textbf{Paige} (\emph{Action}): ...fidgets with scarf/curl...handles envelope carefully...aligns objects with \hl{deliberate care}... \\

\textbf{Paige} (\emph{Dialogue}): "...afraid the questions might change..." \\

\addlinespace[3pt]
\textbf{Round 5:} \\
\textbf{Paige} (\emph{Action}): ...\hl{sets pen down}...moves notebook into shared space...\hl{places both palms} up on the table... \\

\textbf{Paige} (\emph{Dialogue}): "...Find your own way home...maybe I was waiting for someone else to write that part for me." \\

\addlinespace[3pt]
\textbf{Comment:} \\
"Small object-level details help externalize Paige's movement from \hl{waiting and uncertainty} toward \hl{self-reflection and agency}, giving her inner change clearer emotional form." \\

\midrule
\textbf{Model: Qwen3-1.7B} \\
\midrule

\textbf{[1] Partial Emotional Tone Consistency} \\
\addlinespace[2pt]

\textbf{Round 1:} \\
\textbf{Paige} (\emph{Action}): ...stares at the \hl{window}...fingers \hl{hover over her phone}...soft sigh... \\

\textbf{Paige} (\emph{Dialogue}): "...I \hl{hope he's ready by tomorrow}." \\

\addlinespace[3pt]
\textbf{Round 4:} \\
\textbf{Paige} (\emph{Action}): ...fingers hover over her phone...hand placed softly on the table... \\

\textbf{Paige} (\emph{Dialogue}): "...the space between them...now carries the fragile, trembling \hl{weight of words}." \\

\addlinespace[3pt]
\textbf{Comment:} \\
"Despite repetition, Paige's tone remains broadly consistent with an \hl{anxious and anticipatory} character facing an uncertain emotional outcome.
" \\

\midrule

\textbf{[2] Some Use of Restrained Physical Detail to Externalize Anxiety} \\
\addlinespace[2pt]

\textbf{Round 1:} \\
\textbf{Paige} (\emph{Action}): ...\hl{fingers hovering over the phone}...looks out the window... \\

\addlinespace[3pt]
\textbf{Round 6--10:} \\
\textbf{Paige} (\emph{Action}): ...hand placed just below the \hl{glass rim}...gesture quiet and unassuming... \\

\addlinespace[3pt]
\textbf{Comment:} \\
"Repeated gestures around the \hl{phone, window, and table} provide some physically grounded expression of Paige’s anxiety, even if the pattern is overused." \\

\end{longtable}

\normalsize
\subsection{Example of a Complete Interaction Trajectory}

We present the complete interaction trajectory generated by GPT-5.1 when role-playing Benjamin Sullivan as a representative example.

\subsubsection{Scene and Character Information}
\begin{figure}[!ht]
    \centering
    \includegraphics[width=0.8\linewidth]{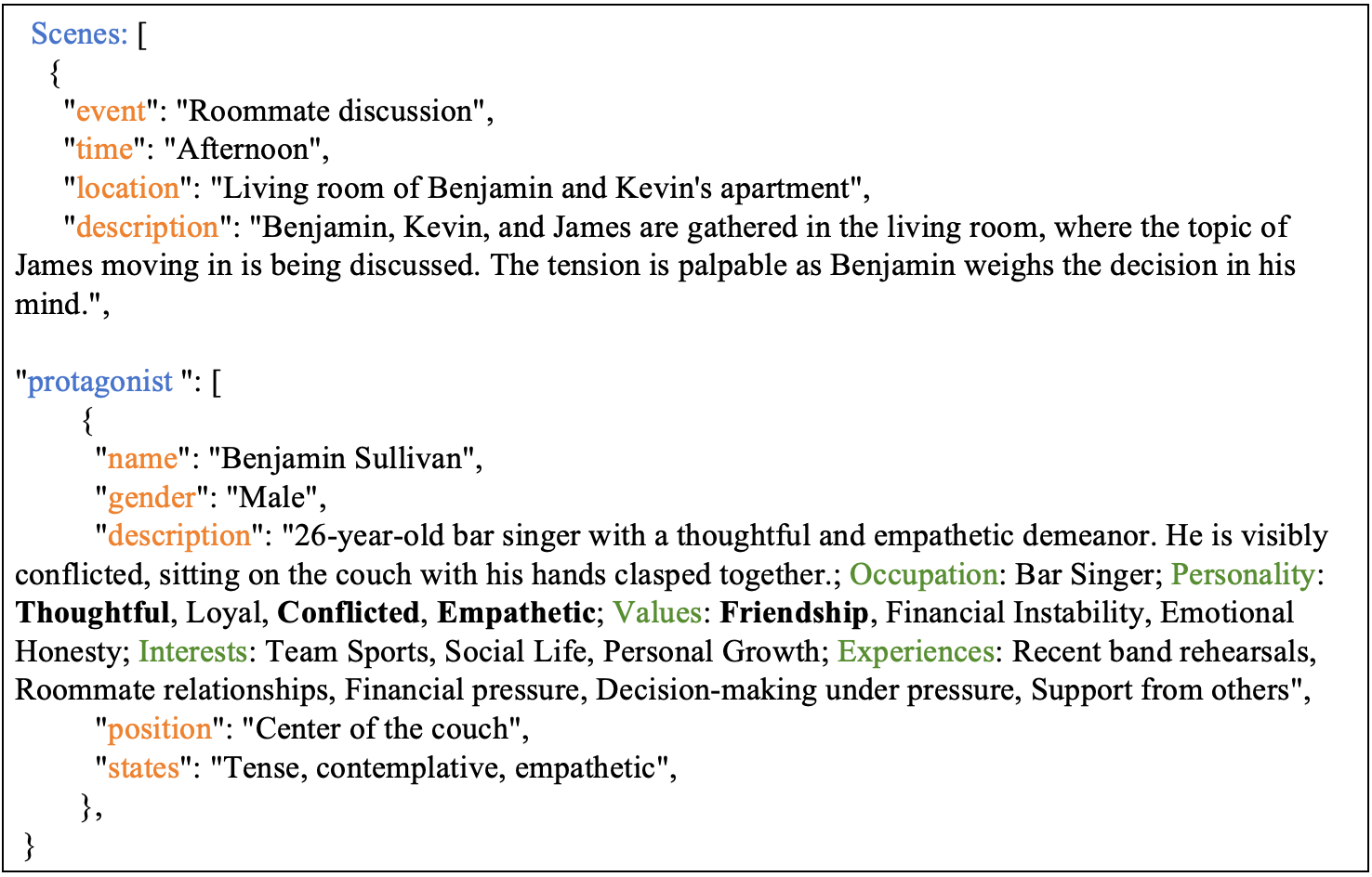}
    \caption{Scene and character information for Benjamin Sullivan.}
    \label{fig:persona_case5}
\end{figure}

\subsubsection{Complete Interaction Trajectory}

\small
\begin{longtable}{@{}p{\dimexpr\linewidth-2\tabcolsep\relax}@{}}
\caption{\textbf{Representative Example of a Complete Interaction Trajectory}: Highlighted phrases (\hl{...}) indicate key interaction information in the trajectory.}
\label{tab:Ben-case-study} \\

\toprule
\textbf{Interaction Trajectory} \\
\midrule
\endfirsthead

\toprule
\endhead

\midrule
\multicolumn{1}{r@{}}{\emph{Continued on next page}} \\
\endfoot

\bottomrule
\endlastfoot

\textbf{Round 1} \\
\midrule

\textbf{Benjamin Sullivan} (\emph{Action}): Benjamin leans forward and slowly rises from the couch, unclasping his hands. He paces a short distance in front of the coffee table, then stops and turns to face both Kevin and James directly. After a brief, steadying breath, he moves to the small side table, picks up the stack of \hl{unopened mail and a folded sheet of paper} with rough budget notes on it, and returns to sit back down between them, placing the papers openly on the table where everyone can see. \\

\textbf{Benjamin Sullivan} (\emph{Dialogue}): Okay, so before we go any further, I need us to actually look at this—between the \hl{bar cutting my Friday shifts} next month and the rent going up \hl{a hundred bucks} in June, the only way this works is if we're all clear on exactly how much you can cover, James, and \hl{what happens if one of us comes up short}. \\

\textbf{Kevin Thompson} (\emph{Action}): Kevin shifts slightly on the couch, his crossed arms tightening for a moment before he reaches over and \hl{grabs the coffee table's notepad, flipping it open to a blank page}. He pauses, eyes scanning the page, then begins scribbling something with the pen resting beside it, his brow furrowed in quiet calculation. After a beat, he slides the notepad toward Benjamin, his hand hovering briefly before letting it settle between them. \\

\textbf{Kevin Thompson} (\emph{Dialogue}): "Look, Ben, I did the numbers again—rent's gonna \hl{eat up 60\% of my paycheck} if we take James in, but if you're on the fence, maybe we could ask him to \hl{cover half the utilities first?}" \\

\textbf{Benjamin Sullivan} (\emph{Reaction} to Kevin Thompson): Benjamin hl{glances down at the notepad}, then slowly reaches out and pulls it closer, his thumb tracing the edge of the page as he exhales, shoulders sinking slightly under the weight of the numbers in front of him. \\

\textbf{Benjamin Sullivan} (\emph{Dialogue} to Kevin Thompson): Kevin, if James \hl{covers half the utilities and we split the extra hundred from June three ways}, that still leaves me short about eighty bucks on months when the bar cancels a gig-so before we say yes or no, can we lay out \hl{a backup plan} for when one of us can’t hit our share? \\

\textbf{James Carter} (\emph{Action}): James Carter steps forward, his hands resting lightly on the back of the couch, and glances briefly between Benjamin and Kevin before pulling his wallet from his pocket and flipping it open. He removes \hl{a folded piece of paper and places it} face-up on the coffee table. \\

\textbf{James Carter} (\emph{Dialogue}): "You can check the numbers yourself—rent, utilities, everything's itemized. \hl{I'm not asking for anything more than what I can afford}, and I promise \hl{I won't let this place slide}." \\

\textbf{Benjamin Sullivan} (\emph{Reaction} to James Carter): Benjamin reaches forward, \hl{carefully unfolds James's paper beside Kevin's notepad}, and lines the two sheets up side by side on the coffee table so all three of them can see the numbers together. \\

\textbf{Benjamin Sullivan} (\emph{Dialogue} to James Carter): James, if we go with this, are you willing to put your name on the lease with us and maybe \hl{set up a shared emergency account-like fifty bucks a month each}-so if my tips tank or Kevin's hours get cut again, we're not one bad week away from losing the apartment? \\

\midrule
\textbf{Round 2} \\
\midrule

\textbf{Benjamin Sullivan} (\emph{Action}): Benjamin shifts forward on the couch and pulls the notepad a few inches closer, then reaches for the pen James set down and neatly \hl{circles two key figures on each document—one on Kevin's budget and one on James's proposal}. He pauses, \hl{comparing them}, then draws a single straight line between the circled totals, creating a shared column of numbers in the empty space between the pages. After a moment, he turns the notepad sideways, \hl{writes a short heading at the top of a fresh section}, and taps that new column twice with the pen, clearly marking it as the starting point for a combined plan the three of them can adjust together. \\

\textbf{Benjamin Sullivan} (\emph{Dialogue}): "If we treat this middle column as our real baseline-what it actually costs the three of us to live here-can we all agree right now to \hl{cap going-out or nonessential spending at}, say, a hundred a month each until we've built at least three months of that total in the emergency account?" \\

\textbf{Kevin Thompson} (\emph{Reaction} to Benjamin Sullivan): Kevin slowly uncrosses his arms and \hl{reaches for the second folded paper-James's detailed expense breakdown}-carefully unfolding it with a deliberate, almost hesitant motion, as if weighing the commitment it represents. \\

\textbf{Kevin Thompson} (\emph{Dialogue} to Benjamin Sullivan): "If \hl{we're capping nonessentials}, what happens if one of us \hl{loses a job or has a medical bill} before the emergency fund's built?" \\

\textbf{James Carter} (\emph{Action}): He flips it open to a blank page and, with the same steady hand that placed his financial proposal on the table, begins to write. His pen moves with quiet purpose, \hl{creating a new list}-this one not of numbers, but of shared responsibilities: groceries, trash day, rent reminders, maintenance checks. \\

\textbf{James Carter} (\emph{Dialogue}): "I also \hl{set aside a buffer for unexpected repairs}-just in case—so we don't have to dip into the emergency fund for a leaky faucet or a broken disposal." \\

\textbf{Benjamin Sullivan} (\emph{Reaction} to James Carter): Benjamin shifts the notepad with the shared responsibilities list a few inches closer to the center of the table, then \hl{adds a small checkbox beside "rent reminders"}. \\

\textbf{Benjamin Sullivan} (\emph{Dialogue} to James Carter): "If we add a line here where we each \hl{write down our actual take-home for the month-and maybe note any big stuff coming up}, like my band's slow season in February-can we agree to check in on this together every four weeks so \hl{none of us ends up quietly drowning?}" \\

\midrule
\textbf{Round 3} \\
\midrule

\textbf{Kevin Thompson} (\emph{Action}): Kevin Thompson slowly \hl{shifting to the fresh column Benjamin had created}. His left hand, still resting on the couch cushion, shifts slightly forward, his posture now more open but still cautious, as if testing the \hl{weight of the compromise being proposed}. \\

\textbf{Kevin Thompson} (\emph{Dialogue}): "Maybe we can start by \hl{splitting the utilities evenly for the next two months} and track how that affects each of our budgets before committing to anything long-term?" \\

\textbf{Benjamin Sullivan} (\emph{Reaction} to Kevin Thompson): Benjamin Sullivan shifts his weight forward and, with a steady but tentative motion, \hl{draws a small asterisk next to Kevin's utilities note} in the shared column, then \hl{underlines "check-in every 4 weeks" once}, reinforcing the trial idea without crossing out any existing numbers. \\

\textbf{Benjamin Sullivan} (\emph{Dialogue} to Kevin Thompson): "Okay, how about this, \hl{for the next two months we split utilities evenly}, and I'll \hl{write down my actual gig income and tips} here each week so you both can see \hl{how tight it gets on my end}, and then when we \hl{do the four-week check-in} we can decide together if we need to tweak the split or cut something from our shared expenses before it blindsides any of us?" \\

\textbf{James Carter} (\emph{Action}): He reaches forward, his fingers hovering briefly \hl{over the list of shared responsibilities}, before extending his hand fully and sliding the notepad toward the center of the table, positioning it directly between the three documents. His movement is deliberate, the pen still in his grip, as if offering it to the group as a shared tool rather than a personal one. \\

\textbf{James Carter} (\emph{Dialogue}): "Let's \hl{start with the buffer}—how much should we allocate for things like the water heater or that flickering light in the hallway?" \\

\textbf{Benjamin Sullivan} (\emph{Reaction} to James Carter): Benjamin Sullivan shifts the pen so it \hl{rests squarely on the shared responsibilities list, then draws a short new line labeled "Emergency Buffer" beneath the existing items}, glancing between Kevin and James as he does. \\

\textbf{Benjamin Sullivan} (\emph{Dialogue} to James Carter): "Maybe we set, like, \hl{thirty bucks each a month into an envelope} in the kitchen drawer as an emergency buffer, and if that starts to feel like too much for any of us-especially on a slow month for my gigs-\hl{we promise to talk about it} before anyone quietly starts skipping other bills to make it work?" \\

\midrule
\textbf{Round 4} \\
\midrule

\textbf{Benjamin Sullivan} (\emph{Action}): Benjamin shifts forward on the couch and \hl{reaches for the envelope box} on the corner of the coffee table, selecting one and placing it in the center of the three documents. He \hl{writes a clear label} across the front, then pulls the notepad slightly closer and adds a short, numbered line beneath "Emergency Buffer," aligning it with the shared responsibilities list. He lightly taps the labeled envelope once, as if \hl{inviting their participation in the new step}. \\

\textbf{Benjamin Sullivan} (\emph{Dialogue}): "If we're all okay with this, we could \hl{start the envelope with ten} each from next month's first paycheck or my first full weekend of gigs, and maybe we write our initials on the back every time we add to it so \hl{it's clear no one's quietly carrying more than they can handle}-does that feel fair to both of you?" \\

\textbf{Kevin Thompson} (\emph{Reaction} to Benjamin Sullivan): Kevin Thompson slowly \hl{picks up the labeled envelope} from the center of the table, turns it over in his hands, and with a careful, deliberate motion, pulls a small black Sharpie from his pocket and \hl{writes his initials on the back}. \\

\textbf{Kevin Thompson} (\emph{Dialogue} to Benjamin Sullivan): "Let's set a reminder for the four-week check-in on my phone calendar so we don't lose track, and maybe we can \hl{add a second buffer line for unexpected rent hikes or service fee spikes?}" \\

\textbf{James Carter} (\emph{Action}): James Carter steps forward, his hand still holding the pen, and reaches out to adjust the position of the folded expense breakdown on the table. He slides it slightly closer to Kevin, \hl{aligning it with the shared responsibilities notepad and Benjamin's budget notes.} \\

\textbf{James Carter} (\emph{Dialogue}): "I've also \hl{added a line for splitting the grocery bill} evenly, but if either of you has a specific dietary need or budget limit, we should note it here so we don't overextend anyone." \\

\end{longtable}

\end{document}